\documentclass[11pt]{article}

\usepackage[preprint]{acl}

\usepackage{times}
\usepackage{latexsym}

\usepackage[T1]{fontenc}

\usepackage[utf8]{inputenc}

\usepackage{microtype}
\usepackage{float}

\usepackage{inconsolata}

\usepackage{graphicx}
\usepackage{subcaption}
\usepackage{multicol}
\usepackage{multirow}

\usepackage{booktabs}    
\usepackage{array}       
\usepackage{makecell}    
\usepackage{footnote}
\usepackage{enumerate}
\usepackage{enumitem}
\usepackage{amsmath}
\usepackage{colortbl}
\definecolor{diffbest}{RGB}{173, 216, 230} 

\usepackage{tcolorbox}
\usepackage{enumitem}

\title{Reliability Challenges in Diffusion Vision–Language Models}

\author{Md. Atabuzzaman \qquad Chris Thomas \\
        Department of Computer Science \\ Virginia Tech \\
        \texttt{\{atabuzzaman, christhomas\}@vt.edu}}

\begin{document}
\maketitle

\begin{abstract}
    Diffusion-based Large Vision-Language Models (dLVLMs) have recently emerged as a compelling alternative to autoregressive (AR) LVLMs, offering advantages in parallel decoding, bidirectional context, and controllable generation. Despite rapid progress, their reliability properties remain largely uncharacterized. We present the first systematic reliability evaluation of hallucination and bias in dLVLMs, benchmarking six diffusion models against 
    competitive AR baselines across four dimensions. Our key findings are: (1) dLVLMs reverse the yes-bias of AR models in binary visual queries; (2) they achieve competitive hallucination rates yet exhibit degraded linguistic 
    quality; (3) they collapse to near-zero accuracy on underrepresented racial groups with opposite-polarity gender bias; and (4) they exhibit accuracy collapse in multiple-choice settings when the correct option is shorter than 
    its distractors, associated with a length prior that emerges at the first denoising step. Tokens committed at late denoising steps with low confidence further correlate with hallucinated content, pointing to a mechanistic signal 
    unique to diffusion generation. These patterns vary across model families, suggesting reliability is shaped by the generative paradigm together with training data.\footnote{\texttt{Code and Dataset: \url{https://github.com/Atabuzzaman/Reliability_dLVLM}}}
\end{abstract}


\section{Introduction}

Large Vision-Language Models (LVLMs) have recently achieved strong performance across diverse vision-language tasks, from visual question answering and mathematical reasoning to document understanding and image captioning~\cite{llava, 
Qwen2.5-VL, internvl2_5}. These models predominantly rely on autoregressive (AR) generation, producing responses token by token in a left-to-right sequence. While effective, this paradigm has fundamental limitations: sequential decoding is hard to parallelize, bidirectional context is unavailable during generation, and AR models are prone to hallucination and systematic output biases~\cite{bang2023multitask, pezeshkpour2024large}.

Discrete diffusion language models have emerged as an alternative to AR generation~\cite{nie2025large, ye2025dream}. Rather than producing tokens sequentially from left to right, these models treat generation as an iterative 
denoising process over discrete tokens, starting from a fully masked sequence 
and progressively unmasking tokens using bidirectional context. This offers 
structural advantages over AR models: parallel decoding enables flexible 
speed-quality trade-offs, bidirectional attention allows conditioning on the 
full response context, and iterative refinement supports constrained generation. 
Recent work has extended discrete diffusion to the multimodal setting~\cite{li2025lavida, 
you2025llada, yu2025dimple, ye2025dreamV, yang2025mmada}, demonstrating that 
dLVLMs achieve competitive or superior performance on standard benchmarks.

Despite this rapid progress, a critical question remains overlooked: \textit{do 
the structural properties of diffusion-based generation alter the reliability of 
LVLMs, specifically their bias and hallucination patterns?} AR models inherit 
well-documented failure modes: they are prone to object hallucination~\cite{li2023evaluating}, 
exhibit strong yes-bias in binary visual queries~\cite{li2023evaluating}, show 
systematic preferences for longer or positionally favored answer options in 
multiple-choice question answering (MCQA) settings~\cite{atabuzzaman2025benchmarking, 
zhao2025large}, and exhibit demographic disparities across gender and racial 
groups~\cite{karkkainen2021fairface, wang2021gender}. Whether dLVLMs share, 
amplify, or alter these failure modes remains largely unexplored.

In this paper, we conduct the first systematic reliability evaluation of dLVLMs, benchmarking six models against competitive AR baselines across four dimensions: (1)~\textbf{object hallucination}, evaluated via the POPE benchmark~\cite{li2023evaluating} across random, popular, and adversarial sampling strategies; (2)~\textbf{open-ended hallucination}, evaluated via the CHAIR metric~\cite{rohrbach2018object} on free-form image captioning; (3)~\textbf{demographic bias}, evaluated via the FairFace dataset~\cite{karkkainen2021fairface} for gender and racial group recognition under two image padding conditions; and (4)~\textbf{selection bias}, evaluated via length-controlled multiple-choice visual questions derived from~\citet{atabuzzaman2025benchmarking}, probing sensitivity to option length as a confounding cue.

Our experiments reveal qualitatively distinct, and in several cases more severe, reliability profiles in dLVLMs than AR counterparts, with patterns varying systematically across model families. These findings establish a new empirical foundation for reliability-aware evaluation of dLVLMs as this model class scales toward deployment. Our main contributions are as follows:

\begin{itemize}[noitemsep, leftmargin=*, topsep=0pt]
    \item Benchmarking dLVLMs against competitive AR baselines across hallucination, demographic bias, and selection bias, providing the first systematic reliability evaluation of this model family.
    \item dLVLMs exhibit systematically different reliability profiles from AR models, associated with the diffusion paradigm, with linguistic quality and 
    hallucination as separable failure modes.
    \item dLVLMs exhibit far more severe length bias in MCQA than AR models, linked to a length preference that appears at the first denoising step.
    \item We identify opposite-polarity gender bias across diffusion model families, near-zero accuracy on underrepresented racial groups, and a 
    denoising confidence trajectory signal correlated with hallucinated content.
\end{itemize}


\section{Related Work}

\textbf{Diffusion Language Models.} Diffusion-based LLMs~\cite{li2022diffusion,nie2025large} 
operate over discrete tokens, treating generation as iterative denoising from a 
fully masked sequence. \citet{nie2025large} proposed LLaDA, a discrete diffusion 
LLM trained from scratch that demonstrates competitive performance with strong AR 
LLMs on in-context learning, instruction following, and reasoning tasks, while 
alleviating limitations such as the reversal curse~\cite{berglund2024reversal}. 
Subsequent work has extended LLaDA along several axes, including long-context 
extrapolation~\cite{liu2025longllada}, sparse Mixture-of-Experts 
scaling~\cite{zhu2025llada}, and AR-based initialization with context-adaptive 
noise scheduling in Dream-7B~\cite{ye2025dream}.


\begin{figure*}[t]
    \centering
    \includegraphics[width=\linewidth]{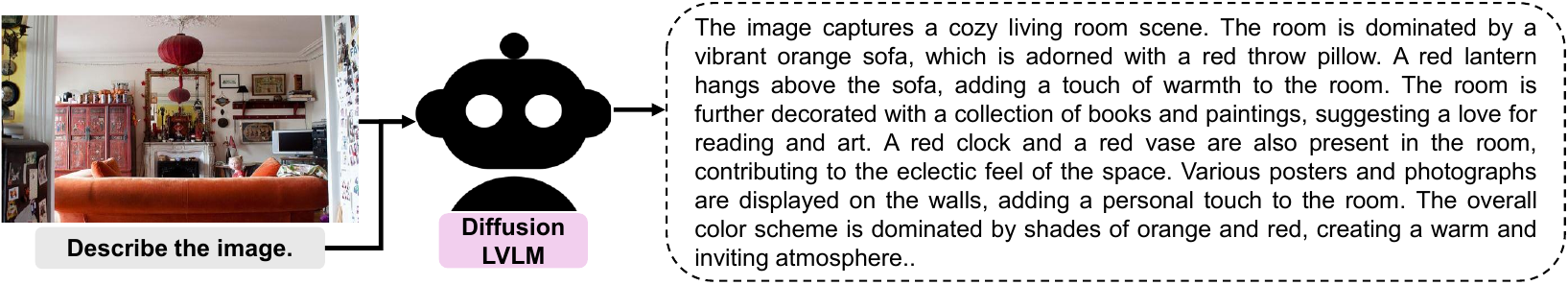}
    \vspace{0.3em}
    \includegraphics[width=\linewidth]{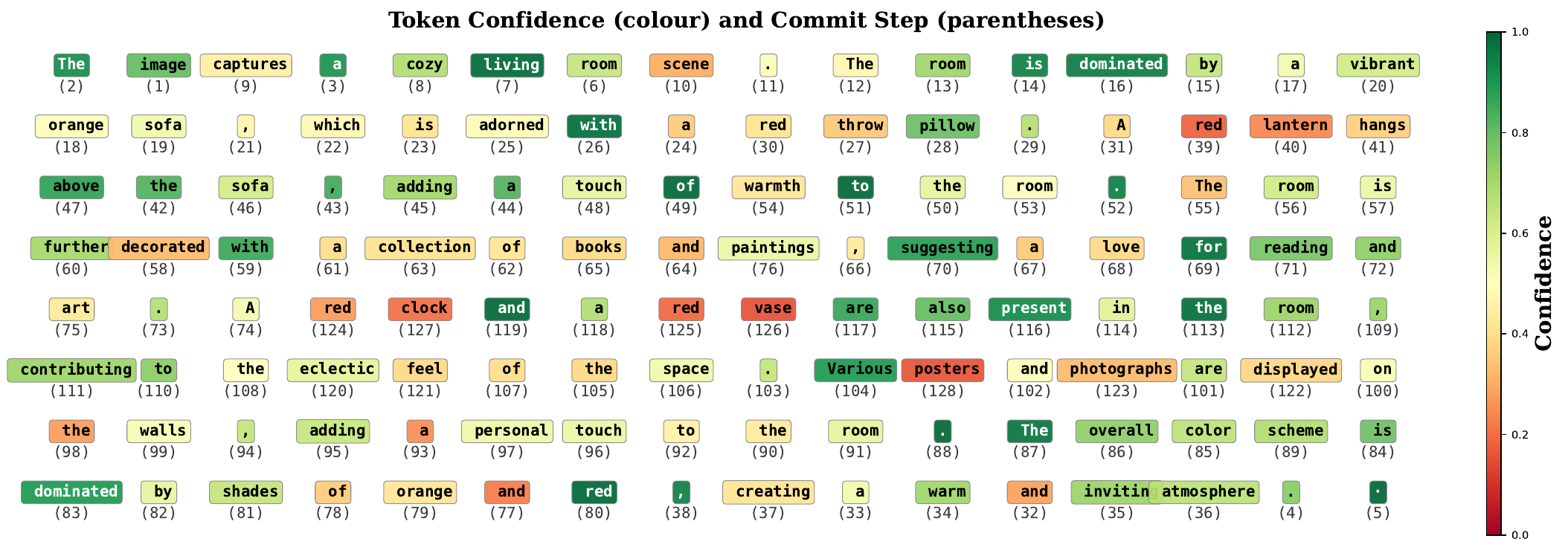}
    \vspace{-2.5em}
    \caption{Qualitative hallucination analysis on a LaViDa-LLaDA caption (MSCOCO).
    \textbf{Top:} The model generates a fluent description of the living room image, but hallucinates objects not present in the scene (\textit{clock}, \textit{vase}, \textit{posters}). 
    \textbf{Bottom:} Per-token confidence (colour) and commit step (parentheses) from the diffusion denoising process. Hallucinated tokens --- \textit{clock} (step~127) and \textit{vase} (step~126) --- and the imprecise \textit{posters} (step~128) are all committed at late denoising steps with low confidence (red), whereas the non-hallucinated \textit{lantern} (step~40), despite low confidence, is committed substantially earlier. This suggests that late commit step combined with low confidence may jointly indicate hallucination risk.}
    \label{fig:qualitative_hallucination}
    \vspace{-1.5em}
\end{figure*}

\noindent \textbf{Diffusion-based LVLMs.} The extension of discrete diffusion models to multimodal settings has recently gained momentum. LaViDa~\citep{li2025lavida} equips discrete diffusion backbones with a vision encoder and jointly fine-tunes for multimodal instruction following, yielding two variants based on the LLaDA and Dream backbones. LLaDA-V~\citep{you2025llada} is a purely diffusion-based LVLM built on LLaDA-8B, demonstrating superior data scalability over an AR baseline. MMaDA~\citep{yang2025mmada} is a unified multimodal diffusion model combining mixed chain-of-thought fine-tuning with a diffusion-specific RL algorithm for text generation, multimodal reasoning, and text-to-image synthesis. Dimple~\citep{yu2025dimple} addresses training instability via an autoregressive-then-diffusion paradigm and confident decoding. Dream-VL~\citep{ye2025dreamV} is built on Dream-7B that demonstrates strong performance.

\noindent \textbf{Hallucination and Bias in LVLMs.} Bias and hallucination in AR LVLMs have been extensively studied. Object hallucination has been benchmarked through POPE~\cite{li2023evaluating}, which probes yes/no object queries across random, popular, and adversarial sampling strategies. Response bias in MCQA settings is well-documented, with models exhibiting systematic preferences for specific answer positions~\cite{pezeshkpour2024large}, verbose options~\cite{zhao2025large}, and particular formatting patterns~\cite{zheng2023judging}; \citet{atabuzzaman2025benchmarking} further demonstrated strong selection bias in fine-grained visual MCQA. Demographic bias has been explored through datasets such as FairFace~\cite{karkkainen2021fairface}, revealing race- and gender-dependent disparities in visual recognition~\cite{wang2021gender}. TraceDet~\cite{chang2025tracedet} leverages intermediate denoising trajectories 
in diffusion LLMs to identify hallucination-relevant steps, achieving improved 
detection accuracy over output-only baselines; however, it targets text-only 
diffusion LLMs and hallucination detection alone. Despite growing interest in 
dLVLMs, their bias and hallucination properties remain largely unexplored. Prior 
work has focused primarily on AR models, leaving open whether diffusion-based 
generation changes these failure modes. To our knowledge, this is the first 
systematic study to address this gap, providing an empirical characterization of 
dLVLM reliability relative to AR counterparts and establishing a foundation for 
reliability-aware evaluation.


\section{Preliminary}

\noindent\textbf{Diffusion-Based Large Vision-Language Models.}
Diffusion-based LVLMs (dLVLMs) treat response generation as an iterative 
denoising process over discrete tokens. The forward process gradually corrupts 
a clean response $\mathbf{x}_0$ into a fully masked sequence $\mathbf{x}_1$ by 
independently replacing tokens with a special \texttt{[MASK]} token. At 
inference, the model starts from a fully masked response and iteratively unmasks 
tokens over $K$ denoising steps. At each step $k$, the model predicts all masked 
positions simultaneously:
\begin{equation}
    p_\theta(\mathbf{x}_0 \mid \mathbf{v}, \mathbf{p}, \mathbf{x}_k) = 
    \prod_{i : x_k^i = \texttt{[M]}} p_\theta(x_0^i \mid \mathbf{v}, \mathbf{p}, 
    \mathbf{x}_k),
\end{equation}
where the denoiser $p_\theta$ uses a full bidirectional attention mask, allowing 
every position to attend to all others. In this work, we evaluate six dLVLMs, 
each built on a discrete masked diffusion backbone but differing in base language 
model, vision encoder integration, and post-training strategy. At each denoising 
step, the model assigns a per-token confidence score (the maximum softmax 
probability over the vocabulary) and selects positions to unmask based on these 
scores. The \textit{commit step} of a token is the step at which it is first 
assigned its final non-mask value. These two signals form the basis of our 
mechanistic hallucination analysis in Section~\ref{sec:hallucination}.


\begin{table}[t]
\centering
\small 
\resizebox{\columnwidth}{!}{%
\begin{tabular}{llcccccc}
\toprule
\textbf{Ca.} & \textbf{Model} &\textbf{Type} & \textbf{Acc.} $\uparrow$ & \textbf{Pre.} $\uparrow$ & \textbf{Rec.} $\uparrow$ & \textbf{F1} $\uparrow$ & \textbf{Yes\%} \\
\midrule

\multirow{13}{*}{\textit{Ra.}}
 & mPLUG-Owl  & AR    & 53.30 & 51.71 & 99.53   & 68.06   & 96.23  \\
 & LLaVA & AR        & 54.43  & 52.32  & \textbf{99.80}  & 68.65  & 95.37  \\
 & LLaVA-Next     & AR & 88.00  & 98.97  & 76.80  & 86.49  & 38.80 \\
 
 & MiniGPT-4      & AR  & 77.83  & 75.38 & 82.67  & 78.86  & 54.83  \\
 & InstructBLIP   & AR  & 88.73  & 85.08 & 93.93  & 89.29 & 55.20  \\
 & Qwen2.5-VL     & AR  & 83.00  & \textbf{100.0}  & 66.00  & 79.52  & 33.00 \\
 & InternVL2.5    & AR  & \textbf{92.60}  & 99.53  & 85.60  & \textbf{92.04}  & 43.00 \\
\cmidrule{2-8}

 & LLaDA-V  & Diff.  & 84.58 & 99.43 & 69.88 & 82.08 & 35.50  \\
 & LaViDa-D & Diff.   & 86.60          & 97.41 & 75.20        & 84.88          & 38.60  \\
 & LaViDa-L  & Diff.   & 84.80          & 98.33 & 70.80        & 82.33          & 36.00  \\
 & MMaDA-M  & Diff.    & 83.00          & 83.40          & \cellcolor{diffbest}82.40           & 82.90          & \cellcolor{diffbest}\textbf{49.40}  \\
 & Dream-VL  & Diff.     & 85.80   & \cellcolor{diffbest}\textbf{100.0}   & 71.60   & 83.45   & 35.80   \\

 & Dimple  & Diff.     & \cellcolor{diffbest}88.00  & 98.97  & 76.80  & \cellcolor{diffbest}86.49  & 38.80     \\

\midrule

\multirow{13}{*}{\textit{Po.}}
 & mPLUG-Owl & AR    & 50.63  & 50.32  & 99.27  & 66.79    & 98.63  \\
 & LLaVA    & AR      & 52.43   & 51.25  & \textbf{99.80}   & 67.72    & 97.37  \\
 & LLaVA-Next     & AR  & 89.80  & \textbf{97.62}  & 81.67  & 88.94  & 42.00 \\

 & MiniGPT-4  & AR    & 68.30          & 64.27          & 82.40           & 72.21          & 64.10  \\
 & InstructBLIP  & AR  & 81.37          & 75.07          & 93.93           & 83.45          & 62.57  \\
 & Qwen2.5-VL  & AR  & 86.40  & 97.41  & 74.90  & 84.68  & 38.60 \\
 & InternVL2.5 & AR  & \textbf{92.40}  & 94.19  & 90.44  & \textbf{92.28}  & \textbf{48.20} \\
\cmidrule{2-8}

 & LLaDA-V       & Diff.  & 87.75 & \cellcolor{diffbest}96.10 & 78.80 & 86.59 & 41.16  \\
 & LaViDa-D  & Diff.  & \cellcolor{diffbest}88.40 & 93.67 & 82.47 & \cellcolor{diffbest}87.71 & \cellcolor{diffbest}44.20  \\
 & LaViDa-L  & Diff.  & 87.60 & 93.55 & 80.88 & 86.75 & 43.40  \\
 & MMaDA-M  & Diff.  & 79.60 & 75.96 & \cellcolor{diffbest}86.85 & 81.04 & 57.40  \\
 & Dream-VL      & Diff.  & 87.00 & 95.15 & 78.09 & 85.78 & 41.20  \\
 & Dimple        & Diff.  & 87.80 & 93.18 & 81.67 & 87.05 & 44.00  \\

\midrule

\multirow{13}{*}{\textit{Ad.}}
 & mPLUG-Owl   & AR   & 50.67  & 50.34   & 99.33   & 66.82    & 98.67  \\
 & LLaVA    & AR      & 50.77  & 50.39   & \textbf{99.87}  & 66.98  & 99.10  \\
 & LLaVA-Next   & AR  & \textbf{86.00}  & 93.20   & 77.42  & 84.58  & 41.20 \\

 & MiniGPT-4  & AR    & 66.60   & 62.45   & 83.27  & 71.37   & 66.67  \\
 & InstructBLIP  & AR   & 74.37   & 67.67    & 93.33   & 78.45  & 68.97  \\
 & Qwen2.5-VL  & AR  & 82.60  & \textbf{96.00}& 67.74  & 79.43  & 35.00 \\
 & InternVL2.5 & AR  & 85.20  & 83.72  & 87.10  & \textbf{85.38}  & \textbf{51.60} \\
\cmidrule{2-8}

 & LLaDA-V  & Diff.  & 82.77 & 92.63  & 70.97 & 80.37 & 38.08  \\
 & LaViDa-D  & Diff.   & 82.60 & 86.76 & 76.61           & 81.37 & 43.80  \\
 & LaViDa-L   & Diff.   & 80.00          & 83.33          & 74.60           & 78.72          & \cellcolor{diffbest}44.40  \\
 & MMaDA-M  & Diff.    & 72.80          & 68.54          & \cellcolor{diffbest}83.47           & 75.27          & 60.40  \\
 & Dream-VL  & Diff.    & \cellcolor{diffbest}{84.20}   & \cellcolor{diffbest}{92.89}   & 73.79  & 82.25  & 39.40   \\
 & Dimple  & Diff.     & 84.00  & 88.18  & 78.23  & \cellcolor{diffbest}{82.91}  & 44.00    \\

\bottomrule
\end{tabular}%
}
\vspace{-0.5em}
\caption{
    POPE benchmark evaluation (MSCOCO). Best overall results per category 
    are in \textbf{bold}; best results among diffusion models are 
    \colorbox{diffbest}{\strut highlighted}. For Yes\%, values closest to 
    50 indicate least response bias. AR = autoregressive; Diff. = 
    diffusion-based; Ca. = category; Acc. = accuracy; Pre. = precision; 
    Rec. = recall; Ra. = random; Po. = popular; Ad. = adversarial.
}
\label{tab:pope_results}
\vspace{-1.5em}
\end{table}


\section{Experiments}

We evaluate six dLVLMs (LLaDA-V~\cite{you2025llada}, LaViDa-LLaDA (LaViDa-L)~\cite{li2025lavida}, LaViDa-Dream (LaViDa-D)~\cite{li2025lavida}, MMaDA-MixCoT (MMaDA-M)~\cite{yang2025mmada}, Dream-VL~\cite{ye2025dreamV}, and Dimple~\cite{yu2025dimple}) against competitive AR baselines across four reliability dimensions. We note that MMaDA-MixCoT corresponds to the publicly released \texttt{MMaDA-8B-MixCoT} checkpoint (Stage 2); the full MMaDA model with UniGRPO RL (Stage 3) was not publicly available at the time of this work~\citep{yang2025mmada}. 

\subsection{Hallucination Evaluation}
\label{sec:hallucination}

We evaluate object hallucination using the POPE benchmark~\cite{li2023evaluating}, which probes yes/no object queries on MSCOCO~\cite{lin2014microsoft} images across three settings: Random (uniform object sampling), Popular (frequently appearing objects), and Adversarial (co-occurring objects designed to elicit hallucination). We report Accuracy, Precision, Recall, F1, and Yes\%, where Yes\% captures the proportion of affirmative responses and serves as a proxy for response bias. We compare against seven AR LVLMs: LLaVA-1.5-7B (LLaVA)~\cite{llava}, LLaVA-1.6-7B (LLaVA-Next)~\cite{liu2024improved}, Qwen2.5-VL-7B-Instruct (Qwen2.5-VL)~\cite{Qwen2.5-VL}, InternVL2.5-8B (InternVL2.5)~\cite{internvl2_5}, mPLUG-Owl (mPLUG)~\cite{ye2023mplug}, MiniGPT-4~\cite{zhu2024minigpt}, and InstructBLIP~\cite{instructblip}. Results are presented in Table~\ref{tab:pope_results}.

\noindent\textbf{dLVLMs Achieve Competitive Hallucination Resistance.} Across all three settings, dLVLMs achieve competitive accuracy relative to AR baselines, with InternVL2.5 retaining the top overall score in each category. Among dLVLMs, Dimple achieves the highest accuracy on Random (88.00\%), LaViDa-Dream on Popular (88.40\%), and Dream-VL on Adversarial (84.20\%). Notably, LaViDa-Dream surpasses InstructBLIP (the strongest among the older AR baselines) by 7.03 and 8.23 percentage points on Popular and Adversarial respectively, demonstrating that dLVLMs have narrowed the gap with competitive AR models on object hallucination.

\noindent\textbf{AR Models Are Yes-Biased; dLVLMs Trend Toward No-Bias.} Older AR models exhibit a pronounced yes-bias, with Yes\% ranging from 54.83\% (MiniGPT-4) to nearly 100\% (mPLUG-Owl, LLaVA), while modern AR models are more calibrated: InternVL2.5 reaches 43.00\% on Random. In contrast, most dLVLMs display the opposite pattern, with Yes\% consistently between 35--45\% across all three settings, yielding high precision (up to 100\% for Dream-VL on Random) but at the cost of reduced recall. The exception is MMaDA-MixCoT, whose Yes\% increases to 60.40\% on Adversarial, suggesting its MixCoT training objective reinforces yes-leaning tendencies. These findings underscore that response bias in dLVLMs is associated with both the generative architecture and the training objective.


\begin{table}[!ht]
\centering
\resizebox{\columnwidth}{!}{%
\begin{tabular}{llccc}
\toprule
\textbf{Type} & \textbf{Model} & \textbf{CHAIR$_\text{I}$ $\downarrow$} & \textbf{CHAIR$_\text{S}$ $\downarrow$} & \textbf{Avg Len} \\
\midrule
\multirow{4}{*}{AR}
 & LLaVA-1.5   & 26.62 & 57.52 & 87.3 \\
 & LLaVA-Next     & 14.50 & 30.14 & 98.8 \\
 & Qwen2.5-VL  & 13.99 & 30.35 & 97.9 \\
 & InternVL2.5 & 12.63 & 27.96 & 84.1 \\
\midrule
\multirow{9}{*}{Diff.}
 & LLaDA-V        & 15.57 & 30.71 & 108.2 \\ 
 & LaViDa-L   & 18.43    & 30.91    & 109.1 \\
 & LaViDa-D   & 15.13    & 29.67   & 92.8  \\
 & Dream-VL       & \textbf{11.69}    & \textbf{23.21}   & 76.4 \\
 & MMaDA-M        & 19.75    & 39.35   & 63.5 \\
 & Dimple         & 19.59    & 34.49   & 99.2 \\
 \cmidrule(lr){2-5}
 &\multicolumn{4}{c}{With step size reduced to half (128 $\rightarrow$ 64)} \\
\cmidrule(lr){2-5}
& LaViDa-L   & 18.87    & 32.57    & 108.0 \\
& Dream-VL       & \textbf{10.06}    & \textbf{18.14}   & 55.5 \\
\bottomrule
\end{tabular}%
}
\vspace{-0.5em}
\caption{CHAIR hallucination evaluation on MSCOCO. Lower is better for CHAIR$_\text{I}$ and CHAIR$_\text{S}$. Avg Len denotes average generated caption length in words. Best results are in bold.}
\label{tab:chair}
\vspace{-1em}
\end{table}


\noindent
\textbf{Open-Ended Hallucination Evaluation.} To complement the binary POPE 
evaluation, we assess hallucination in free-form generation using the CHAIR 
metric~\citep{rohrbach2018object}, which measures how often models mention MSCOCO 
objects not present in the image. We evaluate all models on 500 MSCOCO val2014 
images using the prompt \textit{``Describe the image.''} with 
\texttt{max\_new\_tokens=128} and confidence-based decoding; dLVLMs use 128 
denoising steps. Results report object-level (CHAIR$_\text{I}$) and caption-level 
(CHAIR$_\text{S}$) hallucination rates alongside average caption length, and are 
presented in Table~\ref{tab:chair}.

\noindent\textbf{Some dLVLMs Are Competitive on Open-Ended Hallucination.} Several dLVLMs fall within the competitive range of strong AR baselines: LaViDa-Dream (CHAIR$_\text{I}$: 15.13\%) and LLaDA-V (15.57\%) are comparable to Qwen2.5-VL-7B (13.99\%), while Dream-VL achieves the lowest hallucination rate across all evaluated models (CHAIR$_\text{I}$: 11.69\%, CHAIR$_\text{S}$: 23.21\%), surpassing all AR baselines. Notably, caption length does not explain these results: MMaDA-MixCoT produces the shortest captions (63.5 words) yet the highest CHAIR$_\text{S}$ among dLVLMs (39.35\%), while LLaDA-V and LaViDa-LLaDA generate the longest captions yet maintain moderate hallucination rates, indicating that caption length alone does not determine hallucination risk.

\noindent\textbf{Diffusion Training Introduces Hallucination-Specific Challenges.} 
Dimple shares the same training data~\cite{yu2025dimple} as LLaVA-Next yet 
substantially underperforms it on open-ended hallucination (CHAIR$_\text{I}$: 
19.59\% vs.\ 14.50\%; CHAIR$_\text{S}$: 34.49\% vs.\ 30.14\%), despite comparable 
binary hallucination resistance on POPE (88.00\% vs.\ 88.00\% on Random), 
suggesting diffusion training contributes hallucination-specific challenges 
beyond what is explained by data alone.


\begin{table}[ht]
\centering
\resizebox{\columnwidth}{!}{%
\begin{tabular}{llccccccc}
\toprule
\textbf{Model} & \textbf{Type} & \textbf{Gram.} & \textbf{Rep.} & 
\textbf{Incoher.} & \textbf{Trunc.} & \textbf{Unnat.} & \textbf{OA} & 
\textbf{OA\textsuperscript{$-$T}} \\
\midrule
LLaVA-1.5     & AR    & 0.2 & 0.4 & 0.0 & 19.4 &  0.2 & 20.0 &  0.8 \\
LLaVA-Next    & AR    & 2.2 & 0.0 & 0.0 & 69.2 &  0.4 & 69.2 &  2.6 \\
Qwen2.5-VL    & AR    & 0.2 & 0.2 & 0.0 & 26.0 &  0.2 & \textbf{26.4} &  \textbf{0.6} \\
InternVL2.5   & AR    & 0.2 & 0.4 & 0.0 & 27.0 &  0.6 & 27.6 &  1.0 \\
\midrule
Dimple        & Diff. & 3.0 & 5.2 & 0.2 & 0.2  & 10.2 & 13.8 & 13.8 \\
LLaDA-V       & Diff. & 0.8 & 4.0 & 0.2 & 0.0  &  5.6 &  8.2 &  8.2 \\
LaViDa-L      & Diff. & 1.0 & 2.4 & 0.0 & 0.4  &  1.4 &  3.0 &  3.0 \\
LaViDa-D      & Diff. & 6.4 & 0.8 & 0.0 & 6.0  & 12.0 & 13.0 & 13.0 \\
MMaDA-M       & Diff. & 0.2 & 1.4 & 0.0 & 3.0  &  0.4 &  4.6 &  1.6 \\
Dream-VL      & Diff. & 0.2 & 0.4 & 0.0 & 0.4  &  0.6 &  \textbf{1.6} &  \textbf{1.2} \\
\midrule
\multicolumn{9}{c}{With step size reduced to half (128 $\rightarrow$ 64)} \\
\midrule
LaViDa-L      & Diff. & 60.0 & 64.6 & 11.2 & 14.8 & 82.0 & 88.2 & 87.6 \\
Dream-VL      & Diff. & 21.2 & 21.6 & 0.6  & 7.0  & 29.4 & \textbf{46.4} & \textbf{43.8} \\
\bottomrule
\end{tabular}%
}
\vspace{-0.75em}
\caption{Linguistic quality evaluation on MSCOCO captions using GPT-4o-mini as 
judge. All values are percentages of captions with that error present. 
Overall\textsuperscript{$-$T} denotes overall error rate excluding truncation, 
enabling fair cross-paradigm comparison of intrinsic linguistic quality. Gram.\ = 
grammatical error; Rep.\ = repetition; Incoher.\ = incoherence; Trunc.\ = 
truncation; Unnat.\ = unnatural phrasing; OA \ = Overall.}
\label{tab:linguistic_quality}
\vspace{-1em}
\end{table}


\noindent\textbf{Linguistic Quality Evaluation.} To assess whether CHAIR results 
are confounded by surface-level generation quality, we evaluate all generated 
captions using GPT-4o-mini as a judge across five error types: grammatical error, 
repetition, incoherence, truncation, and unnatural phrasing, using deterministic 
decoding. The prompt is provided in Appendix~\ref{sec:linguistic_prompt}. Results 
are presented in Table~\ref{tab:linguistic_quality}.

Unnatural phrasing is the primary dLVLM-specific linguistic failure, though not 
universal: MMaDA-M (0.4\%), LaViDa-LLaDA (1.4\%), and Dream-VL (0.6\%) are 
competitive with AR baselines (0.2--0.6\%), while LaViDa-Dream (6.4\% grammatical 
error, 12.0\% unnatural phrasing) and Dimple (3.0\%, 10.2\%) show elevated rates. 
Incoherence is negligible across all models (0.0--0.2\%). Truncation reveals a 
complementary limitation of AR decoding under fixed budgets: LLaVA-Next truncates 
69.2\% of captions and InternVL2.5 27.0\% under \texttt{max\_new\_tokens=128}, 
while dLVLMs rarely truncate (0.0--6.0\%). We report Overall\textsuperscript{$-$T} 
for cross-paradigm comparison: AR models achieve 0.6--2.6\% and dLVLMs range from 
1.2\% (Dream-VL) to 13.8\% (Dimple).


\noindent\textbf{Effect of Denoising Steps on Generation Quality.} To examine 
how denoising steps affect hallucination and linguistic quality, we evaluate 
LaViDa-LLaDA and Dream-VL with step size reduced to half (128 vs.\ 64). Results 
are shown in Tables~\ref{tab:chair} and~\ref{tab:linguistic_quality}. 
\textbf{Reducing steps causes dramatic linguistic degradation}: LaViDa-LLaDA's 
overall error rate rises from 3.0\% to 87.6\% (grammatical: 1.0\%→60.0\%; 
repetition: 2.4\%→64.6\%), while Dream-VL also degrades substantially (1.2\% 
to 43.8\%). Hallucination rates are less sensitive: LaViDa-LLaDA increases 
marginally (CHAIR\textsubscript{I}: 18.43\%→18.87\%), and Dream-VL's apparent 
improvement (11.69\%→10.06\%) is accompanied by a substantial caption length 
reduction (76.4→55.5 words), suggesting shorter outputs rather than improved 
grounding. These results confirm that reducing denoising steps primarily degrades 
linguistic quality, while hallucination reflects grounding properties established 
during training.


\begin{table}[ht]
\centering
\resizebox{\columnwidth}{!}{%
\begin{tabular}{llccc}
\toprule
\textbf{Model} & \textbf{Decoding} & \textbf{CHAIR\textsubscript{I}~$\downarrow$} & 
\textbf{CHAIR\textsubscript{S}~$\downarrow$} & \textbf{Ling. Error~$\downarrow$} \\
\midrule
Dimple       & Diffusion & 19.59 & 34.49 & 13.8\% \\
Dimple       & AR-style         & \textbf{16.23} & \textbf{31.60} &  \textbf{2.0}\% \\
\midrule
LaViDa-L & Diffusion & 18.43 & 30.91 &  3.0\% \\
LaViDa-L & AR-style         & \textbf{17.50} & \textbf{30.23} &  \textbf{1.2}\% \\
\bottomrule
\end{tabular}%
}
\vspace{-0.75em}
\caption{CHAIR hallucination and overall linguistic error rate (excluding truncation) under default diffusion vs.\ AR-style decoding, using the same model weights with no retraining. Ling.\ Error = overall linguistic error rate excluding truncation.}
\label{tab:ar_chair}
\vspace{-1em}
\end{table}

\noindent\textbf{AR-Style Decoding Ablation.} To investigate whether the diffusion commitment order independently drives hallucination and linguistic quality, we apply AR-style decoding (left-to-right, one token per step) to Dimple and LaViDa-LLaDA using the same model weights with no retraining, isolating the decoding paradigm while holding architecture, training data, and vision encoder constant. Results are presented in Table~\ref{tab:ar_chair}.

\textbf{AR-style decoding substantially improves linguistic quality.} Dimple's 
error rate drops from 13.8\% to 2.0\%, within the AR baseline range (0.6--2.6\%); 
LaViDa-LLaDA similarly improves from 3.0\% to 1.2\%, matching the best AR 
baselines. However, hallucination persists: CHAIR\textsubscript{I} improves only 
modestly from 19.59\% to 16.23\% for Dimple and from 18.43\% to 17.50\% for 
LaViDa-LLaDA, both remaining substantially above InternVL2.5-8B (12.63\%) and 
Dream-VL (11.69\%). The pattern is consistent across both models: decoding order 
governs surface fluency, while hallucination rates are largely invariant to it. 
This dissociation suggests hallucination reflects grounding-related challenges 
rather than generation artifacts, consistent with Dream-VL's low hallucination 
reflecting better visual grounding despite its diffusion backbone.

\noindent\textbf{Qualitative Analysis.} Figure~\ref{fig:qualitative_hallucination} 
shows an illustrative example of per-token confidence and commit step for a 
LaViDa-LLaDA caption. Hallucinated tokens \textit{clock} (step 127) and 
\textit{vase} (step 126) are committed at late denoising steps with low 
confidence, as is \textit{posters} (step 128), the latest and least confident 
token in the sequence. In contrast, the grounded \textit{lantern} (step 40), 
despite similarly low confidence, is committed substantially earlier. This 
suggests that late commit step combined with low confidence may jointly indicate 
hallucination risk in dLVLMs. Appendix~\ref{sec:qualitative_app} provides 
additional examples across distinct image types.


\noindent\textbf{Quantitative Validation of the Commit-Step Signal.} To move beyond the qualitative pattern in Figures~\ref{fig:qualitative_hallucination}, \ref{fig:qualitative_hallucination_lavida_dream} and \ref{fig:qualitative_hallucination_app}, we quantitatively validate the commit-step hallucination signal using the CHAIR annotations from Section~\ref{sec:hallucination}. For each generated caption we treat CHAIR-flagged hallucinated object tokens as positives and grounded object tokens as negatives, and evaluate commit step and per-token confidence, individually and jointly via a 5-fold cross-validated logistic regression, as hallucination predictors, reporting ROC-AUC and PR-AUC over the 500-image MSCOCO caption set across two backbones (LaViDa-LLaDA and LaViDa-Dream).

\begin{table*}[t]
\centering
\small
\begin{tabular}{lcccccccc}
\toprule
Model & Obj. & Base & Step (H/G) & Conf. (H/G) & AUC$_{\text{step}}$ & AUC$_{\text{conf}}$ & AUC$_{\text{joint}}$ & PR-AUC$_{\text{step}}$ \\
\midrule
LaViDa-L & 968 & 0.190 & 63.8 / 37.6 & 0.533 / 0.615 & \textbf{0.699} & 0.610 & \textbf{0.688} & \textbf{0.374} \\
LaViDa-D & 999 & 0.152 & 64.3 / 42.2 & 0.579 / 0.685 & 0.667 & \textbf{0.626} & 0.673 & 0.261 \\
\bottomrule
\end{tabular}
\vspace{-0.75em}
\caption{Commit step and confidence as hallucination predictors on CHAIR object tokens. H/G denotes mean commit step and mean confidence for hallucinated vs. grounded tokens. Higher commit step and lower confidence indicate hallucination; scores are oriented accordingly. Best results are in bold.}
\label{tab:commit-step-auc}
\vspace{-1em}
\end{table*}

Commit step and confidence separate hallucinated from grounded object tokens consistently across both backbones (Table~\ref{tab:commit-step-auc}). Hallucinated objects commit substantially later, a 22 to 26 step gap on a 128-step budget, and with lower confidence, yielding above-chance ROC-AUC on both models, with commit step the stronger single feature (0.699 and 0.667). PR-AUC for commit step reaches 0.374 and 0.261 respectively, roughly double the corresponding base rate (0.190 and 0.152), confirming the separation survives class imbalance. The two features are correlated, since late commits tend to be low-confidence, which explains why the joint model offers little gain over commit step alone. We treat this as evidence of a hallucination-risk pattern specific to diffusion generation, clearly distinguishable from standard autoregressive decoding, since AR generation has no analogous commit-step trajectory at any stage.

\noindent
\textbf{Attention-Based Analysis.} To further investigate the grounding failure 
hypothesis, we extracted attention weights from answer token positions to image 
patch tokens in LaViDa-LLaDA across three representative layers (8, 16, 24), 
examining mean attention, peak attention, and entropy for hallucinated versus 
grounded tokens at their specific commit steps. We found no statistically 
distinguishable difference between the two categories across any layer or 
aggregation strategy: mean attention values were nearly identical (0.0005 
vs.\ 0.0005) and entropy showed no consistent pattern (4.41 vs.\ 4.40). This 
null result does not rule out grounding failures as a contributing factor; 
attention weights aggregated over all image patches may be too coarse to capture 
the spatially localized or cross-layer grounding signals that distinguish 
hallucinated from grounded tokens. The commit-step and confidence trajectory 
remain the more discriminative signals in this setting. Details are provided in 
Appendix~\ref{sec:attention_analysis}.


\subsection{Demographic Bias Evaluation}
\label{sec:demo_bias_main}

We evaluate all models on gender classification and racial recognition using the 
FairFace dataset~\cite{karkkainen2021fairface}, which provides balanced coverage 
across seven racial groups (Black, East Asian, Indian, Latino Hispanic, Middle 
Eastern, Southeast Asian, and White) and two genders. We construct a balanced 
subset of 2,000 images from the FairFace validation split (142 images per race 
$\times$ gender group) under two face-crop padding settings: padding 0.25 (tight 
crop, isolating facial features) and padding 1.25 (loose crop, retaining background context). The evaluation protocol is provided in Appendix~\ref{sec:fairface_protocol}. 
Full results including gender and per-race accuracy, the female--male accuracy gap, per-model accuracy change across padding settings ($\Delta$), and systematic 
misclassification patterns are provided in 
Appendices~\ref{sec:fairface_results}--\ref{sec:demographic_confusion}.

\noindent\textbf{Output Validity.} To rule out evaluation artifacts, we verify that near-zero subgroup accuracies reflect genuine misclassification rather than parsing failure. Across all diffusion models, every prediction at both padding settings parses to a valid FairFace label (0 of 1{,}988 non-matching outputs per run, for every model), so no accuracy is lost to unparseable output. Table~\ref{tab:race_confusion} confirms these are confident misclassifications into other valid categories rather than malformed responses: MMaDA-MixCoT's 0.00\% on Latino Hispanic and Southeast Asian at padding 0.25 consists entirely of predictions distributed across other valid race labels (Latino Hispanic to Indian 35\%, White 26\%; Southeast Asian to White 33\%, Indian 30\%).


\noindent\textbf{Overall Race Recognition Performance.} AR models consistently outperform dLVLMs on race recognition. InternVL2.5-8B achieves the highest overall accuracy at padding 1.25 (69.32\%), followed by Qwen2.5-VL-7B (68.41\%). Most dLVLMs lag significantly behind, with LaViDa-LLaDA reaching 50.65\% at best and MMaDA-MixCoT barely exceeding 23\%. The exception is Dream-VL (68.11\%), approaching AR-level performance, suggesting its stronger backbone yields richer racial feature representations.

\noindent\textbf{Systematic Failure on Underrepresented Racial Groups.} All 
models struggle with Latino Hispanic and Southeast Asian categories. MMaDA-MixCoT 
achieves 0\% on both groups at padding 0.25, while LaViDa-LLaDA reaches only 
0.35\% and 1.06\% respectively. Table~\ref{tab:race_confusion} reveals systematic 
misclassification into more frequent groups: Latino Hispanic faces are misclassified 
as White or Indian, while Southeast Asian faces are predominantly predicted as East 
Asian (73--84\% in LLaVA-1.5-7B and LaViDa-Dream), suggesting a shared bias 
pattern across both AR and diffusion models.

\noindent\textbf{Context Sensitivity and Divergent Backbone Patterns.} Loose cropping consistently improves performance, with dLVLMs exhibiting larger sensitivity than AR models: LaViDa-LLaDA improves by +10.41\% and MMaDA-MixCoT 
by +13.73\% overall, with MMaDA-MixCoT gaining +24.30\% on East Asian from 
context alone, suggesting substantial reliance on background cues for this group. 
The two LaViDa variants further reveal backbone-specific prototype hierarchies 
despite sharing the same training pipeline: LaViDa-Dream displays a strong East 
Asian dominance bias (63\% of Latino Hispanic, 84\% of Southeast Asian), while 
LaViDa-LLaDA defaults to Indian and White, achieving higher Indian accuracy 
(64.08\% vs.\ 25.35\%). These divergent misclassification targets, arising from 
models that share the training pipeline and vision encoder but differ in backbone 
(Dream-7B vs.\ LLaDA-8B), are consistent with the diffusion backbone influencing 
which racial prototypes dominate predictions.

\noindent\textbf{Gender Bias: Opposite Polarity Across Model Families.} Gender 
classification reveals a divergence in bias polarity at tight crop. AR models 
exhibit small, stable gaps (LLaVA-1.5-7B: $-$2.71; InternVL2.5-8B: $-$1.00; 
Qwen2.5-VL-7B: +3.22), while dLVLMs exhibit much larger and more variable gaps. 
LaViDa-Dream and LaViDa-LLaDA strongly favor females (Gap = +13.48 and +10.26), 
while MMaDA-MixCoT misclassifies females more often (Gap = $-$23.34), with female 
accuracy at 68.51\% versus 91.85\% for males. This asymmetry nearly vanishes at 
loose crop (Gap = +0.10), and LaViDa-Dream's gap narrows from +13.48 to +5.84 
and LaViDa-LLaDA's from +10.26 to +3.73, suggesting dLVLMs encode insufficient 
discriminative signal for female facial features without contextual cues. AR 
models show negligible padding sensitivity ($\Delta < 4\%$), indicating more 
robust gender representations.

\noindent\textbf{Label Ambiguity and Its Limits.} FairFace labels reflect annotator-perceived race, which is inherently ambiguous for several categories, notably Latino Hispanic and Southeast Asian, both of which overlap visually with neighboring groups. Accuracy figures in this evaluation should accordingly be read as agreement with FairFace's perceived-race annotations rather than as ground-truth identity. This ambiguity, however, does not account for the scale of the failures observed: the systematic, backbone-dependent absorption patterns in Table~\ref{tab:race_confusion} (e.g., LaViDa-Dream routing 63--84\% of Latino Hispanic and Southeast Asian faces into East Asian, versus LaViDa-LLaDA routing the same groups into Indian and White) are model-specific and directionally inconsistent with a shared annotation artifact, indicating the failures reflect genuine model bias beyond label ambiguity.


\begin{table}[!ht]
\centering
\small
\resizebox{\columnwidth}{!}{%
\begin{tabular}{ll|cccc}
\toprule
\textbf{Model} &
\makecell{\textbf{Option}\\\textbf{Order}} &
\makecell{\textbf{Equal}\\\textbf{Long}} &
\makecell{\textbf{Equal}\\\textbf{Short}} &
\makecell{\textbf{Shorter}\\\textbf{Correct}} &
\makecell{\textbf{Longer}\\\textbf{Correct}} \\
\midrule
\multicolumn{6}{l}{\textbf{CUB With Class Name}} \\
\addlinespace[2pt]

LLaDA-V & ABCD  & 60.70  & 60.10  & 37.40  & 92.50  \\
\addlinespace[2pt]

\multirow{2}{*}{LaViDa-L} & ABCD   & 47.40 & 48.60 & 11.50 & \textbf{96.60} \\
& DCBA    & 45.70 & 48.80 & 13.40 & 94.80 \\ 
\addlinespace[2pt]

LaViDa-D &ABCD     & 45.10  & 46.50  & 9.00  & 92.60 \\
\addlinespace[2pt]

MMaDA-M & ABCD     & 37.40  & 40.80  & 11.40  & 91.70 \\
\addlinespace[2pt]

Dream-VL & ABCD    & \textbf{68.20}  & \textbf{68.40}  & \textbf{42.10}  & 93.00 \\
\addlinespace[2pt]
Dimple & ABCD     & 54.30  & 54.70  & 26.80  & 87.10 \\

\cmidrule(lr){1-6}
\multicolumn{6}{l}{\textbf{CUB Without Class Name}} \\
\addlinespace[2pt]

LLaDA-V & ABCD & \textbf{63.60}  & 51.30  & \textbf{47.00}  & 82.30 \\ 
\addlinespace[2pt]

\multirow{2}{*}{LaViDa-L} & ABCD  & 51.90 & 48.10 &  6.40 & \textbf{97.70} \\ 
& DCBA   & 52.30 & 45.50 & 9.40  & 97.20 \\ 
\addlinespace[2pt]

LaViDa-D & ABCD   & 38.60  & 48.90  & 2.80  & 96.50  \\
\addlinespace[2pt]

MMaDA-M & ABCD   & 39.00  & 42.10  & 12.80  & 89.70  \\
\addlinespace[2pt]

Dream-VL & ABCD   & 63.00  & \textbf{57.90}  & 30.70   & 84.40  \\ 
\addlinespace[2pt]

Dimple & ABCD  & 50.00  & 49.90  & 19.70  & 80.90  \\  

 \cmidrule(lr){1-6}


\multicolumn{6}{l}{\textbf{Dog With Class Name}} \\
\addlinespace[2pt]

\multirow{2}{*}{LLaDA-V}  & ABCD  & 57.83  & 55.33  & 45.50   & 82.67 \\
  & DCBA  & 55.67 & 54.00 & 44.83  & 82.67 \\
  \addlinespace[2pt]
  
LaViDa-L &ABCD    &  48.50  & 44.33  & 21.67  & 92.67   \\
\addlinespace[2pt]

\multirow{2}{*}{LaViDa-D} & ABCD   & 59.00  & 52.50  & 21.50  & 91.67 \\
  & DCBA  & 59.00  & 54.67  & 22.50  & \textbf{92.83} \\
  \addlinespace[2pt]

\multirow{2}{*}{MMaDA-M} & ABCD   & 37.33  & 37.33  & 24.50  & 72.17 \\
& DCBA    & 41.17  & 37.83  & 29.33  & 67.67 \\
\addlinespace[2pt]

\multirow{2}{*}{Dream-VL}
& ABCD    & \textbf{72.33}  & \textbf{66.50}  & \textbf{56.00}  & 87.33  \\
& DCBA    & 71.33  & 66.33  & 55.00  & 85.50  \\
  
\cmidrule(lr){1-6}
\multicolumn{6}{l}{\textbf{Dog Without Class Name}} \\
\addlinespace[2pt]

\multirow{2}{*}{LLaDA-V} & ABCD & 54.17 & 51.17  & 33.00  & 86.67\\ 
 & DCBA & 50.17  & 51.67  & 28.67  & 84.17 \\ 
\addlinespace[2pt]

LaViDa-L & ABCD & 45.67  & 45.83  & 5.33  & 98.83  \\ 
\addlinespace[2pt]

\multirow{2}{*}{LaViDa-D} & ABCD  & 48.17  & 54.83  & 4.00  & 98.17  \\ 
 & DCBA  & 45.50  & 52.17  & 3.17  & \textbf{99.00} \\ 
 \addlinespace[2pt]

\multirow{2}{*}{MMaDA-M} & ABCD  & 38.50  & 39.67  & 20.83  & 81.00 \\ 
& DCBA  & 33.67  & 38.17  & 11.83  & 81.00  \\ 

\multirow{2}{*}{Dream-VL} & ABCD  & \textbf{62.17}  & \textbf{59.67}  & \textbf{34.17}  & 88.17  \\ 
& DCBA  & 56.33  & \textbf{59.67}  & 26.33  & 89.67  \\ 

\cmidrule(lr){1-6}
\multicolumn{6}{l}{\textbf{Dog (Easy) With Class Name}} \\
\addlinespace[2pt]

\multirow{2}{*}{LLaDA-V}
& ABCD    & \textbf{100.00}  & \textbf{100.00}  & \textbf{100.00}  & \textbf{100.00}  \\ 
& DCBA   & 99.83  & \textbf{100.00}  & \textbf{100.00}  & \textbf{100.00}  \\
\addlinespace[2pt]

\multirow{2}{*}{LaViDa-L}
& ABCD    & \textbf{100.00}  & \textbf{100.00}  & \textbf{100.00}  & \textbf{100.00}  \\ 
& DCBA  & \textbf{100.00}  & \textbf{100.00}  & \textbf{100.00}  & \textbf{100.00}  \\ 
\addlinespace[2pt]

\multirow{2}{*}{LaViDa-D}
& ABCD    & 99.17  & 99.17  & 97.50  &    \textbf{100.00}   \\ 
& DCBA   & 99.17   & \textbf{100.00}  & 98.33  & \textbf{100.00} \\ 
\addlinespace[2pt]

\multirow{2}{*}{MMaDA-M}
& ABCD   & 88.33  & 89.83  & 79.33  & 96.83 \\ 
& DCBA   & 87.17  & 89.00  & 76.67  & 96.17  \\
\addlinespace[2pt]

Dimple & ABCD  & \textbf{100.00} & \textbf{100.00}  & 99.50  & \textbf{100.00}  \\

\bottomrule
\end{tabular}%
}
\vspace{-0.5em}
\caption{Performance of diffusion LVLMs in Accuracy (\%) across option-length conditions and option-order formats. Best results are in bold.}
\label{tab:mcqa_length_bias_results}
\vspace{-1em}
\end{table}


\subsection{Selection Bias Evaluation}
To evaluate selection bias in dLVLMs, we use hard-difficulty MCQs from the fine-grained visual MCQA benchmark of \citet{atabuzzaman2025benchmarking}, covering two domains: CUB-200-2011 (200 bird species)~\cite{wah2011caltech} and Stanford Dogs (120 dog breeds)~\cite{khosla2011novel}. Using GPT-4o~\cite{gpt4o}, we generate four length-controlled variants per question while preserving semantic content: \textbf{Equal Long}, \textbf{Equal Short}, \textbf{Shorter Correct} (correct answer short, distractors long), and \textbf{Longer Correct} (correct answer long, distractors short). The latter two conditions directly probe length bias by placing the correct answer at a systematic length disadvantage or advantage, while Equal Long and Equal Short serve as an internal validity check: accuracy differs by roughly 1 to 5 points in most cases, with some larger exceptions (e.g., LLaDA-V and LaViDa-Dream on CUB without class names), indicating that GPT-4o's rewrites largely preserve the semantic content needed to answer and that length, not precision loss, is the primary driver of the length-bias gap. We evaluate all models with and without explicit class names in the answer options, as class names partially mitigate length bias by providing semantic anchors~\cite{atabuzzaman2025benchmarking}, and under both standard (ABCD) and reversed (DCBA) option orders to control for positional bias. A full comparison with AR baselines is provided in Table~\ref{tab:cub_diffusion_vs_ar} (Appendix~\ref{sec:selection_comp}).


\noindent\textbf{Diffusion LVLMs Exhibit Extreme Length Bias.} Table~\ref{tab:mcqa_length_bias_results} reveals that most dLVLMs suffer from severe length bias across datasets. In the Shorter Correct condition, accuracy collapses dramatically: LaViDa-Dream drops to 2.80\% on CUB without class names, LaViDa-LLaDA to 6.40\%, and MMaDA to 12.80\%. Conversely, all models achieve 68--99\% in the Longer Correct condition, indicating that dLVLMs overwhelmingly select longer options when visual discrimination is hard. The resulting bias gap exceeds 85 percentage points for LaViDa-LLaDA and LaViDa-Dream without class names. LLaDA-V is a notable exception (37.40\% on CUB Shorter Correct with class names, 47.00\% without), suggesting its backbone confers greater robustness to length cues. Dimple also shows reduced but non-negligible bias (26.80\% Shorter Correct on CUB with class names), indicating its AR-then-diffusion training provides partial mitigation.

\noindent\textbf{AR LVLMs Are Substantially More Robust.} AR models maintain meaningfully higher accuracy under the Shorter Correct condition, typically achieving 19--51\% on CUB and 16--59\% on Dogs with class names. Qwen2.5-VL-7B achieves 50.70\% on CUB Shorter Correct with class names, more than four times higher than most dLVLMs, though the gap narrows for weaker AR backbones. This suggests AR models integrate visual evidence more reliably against misleading length cues, while most dLVLMs default to length as a dominant heuristic.

\noindent\textbf{Class Names Reduce But Do Not Eliminate Bias.} Class names provide modest improvement on Dogs (e.g., LaViDa-LLaDA Shorter Correct improves from 5.33\% to 21.67\%) but negligible benefit on CUB, where LaViDa-Dream's Shorter Correct accuracy remains at 9.00\% with class names versus 2.80\% without. AR models benefit more consistently, suggesting dLVLMs lack the semantic grounding needed to leverage class information against competing length signals.

\noindent\textbf{Length Bias Emerges Under Visual Uncertainty.} Comparing ABCD and DCBA orderings reveals differences in Shorter Correct and Longer Correct accuracy that are small in most cases (typically within 0--3 percentage points), though a few conditions show larger gaps (up to 9 points, e.g., MMaDA-MixCoT and Dream-VL on Dogs without class names), indicating that option length, rather than position, is the primary driver of the bias. The Dog (Easy) results further support this: most dLVLMs achieve high accuracy when distractors come from dissimilar categories (87--100\%, with MMaDA-MixCoT lower at 76--97\%), but performance collapses under fine-grained conditions (Hard). This suggests dLVLMs fall back on length heuristics when visual evidence is insufficient, a pattern that mirrors AR model findings~\cite{atabuzzaman2025benchmarking} but is far more pronounced.

\noindent
\textbf{Mechanistic Analysis of Length Bias.} To investigate why dLVLMs exhibit such extreme length bias, we analyze the denoising trajectory of LaViDa-LLaDA on the Shorter Correct and Longer Correct conditions of CUB (1,000 examples each). MCQA responses are short (a single letter A--D), so we use 8 denoising steps over a response buffer of 8 tokens; padding positions are the non-answer slots in this buffer, which the model fills before committing the answer letter. In the Longer Correct condition, the model achieves 97.70\% accuracy with mean confidence 0.96 at step~0, and the step-0 prediction matches the final answer in 99.80\% of cases. In the Shorter Correct condition, accuracy collapses to 6.40\%, while the step-0 prediction still matches the final answer in 97.30\% of cases, indicating that denoising rarely revises the model's initial choice. Since final accuracy is only 6.40\%, at least 90.9\% of examples (909 of 1,000) have a step-0 prediction that already selects the longer distractor and is never revised by any subsequent denoising step. The mean commit step for the answer position is nonetheless the final step in both conditions, despite the prediction being effectively determined at step~0. These findings reveal that length bias is a one-shot prior: in the vast majority of cases, the step-0 prediction already favors the longer option, and denoising does not revise it.


\section{Conclusion}
We presented the first systematic reliability evaluation of dLVLMs, benchmarking six diffusion models against competitive AR baselines across four reliability dimensions. Our results reveal qualitatively distinct and, in several cases, more severe reliability challenges than AR counterparts: dLVLMs trend toward a no-bias pattern on binary visual queries; they achieve competitive open-ended hallucination yet exhibit degraded linguistic quality in several backbones; they collapse to 
near-zero accuracy on some underrepresented racial groups with opposite-polarity gender bias; and they exhibit dramatically larger length-bias gaps in MCQA, associated with a one-shot length prior established at the first denoising step. Late-committed, low-confidence tokens show a consistent pattern with hallucinated content, a signal unique to diffusion generation, which we quantitatively validate across two backbones (ROC-AUC up to 0.699 on CHAIR object tokens). These patterns vary across diffusion backbones, associated with the generative paradigm as a contributing factor alongside training data and scale. We hope these findings motivate reliability-aware evaluation and training strategies for diffusion-based multimodal generation.


\section*{Limitations}

Our study provides the first systematic reliability evaluation of dLVLMs, and we acknowledge several limitations. First, while we conduct controlled comparisons (Dimple vs.\ LLaVA-Next sharing identical training data, AR-style decoding ablations on Dimple and LaViDa-LLaDA, and backbone comparisons within the LaViDa pipeline), full isolation of the generative mechanism from confounds such as vision encoders and model scale requires more extensive ablations. Second, our linguistic quality evaluation relies on GPT-4o-mini as judge and has not been 
validated against human annotations. Third, our evaluation covers four reliability dimensions and does not extend to other important failure modes such as toxicity, sycophancy, or temporal reasoning. We hope this framework serves as a foundation for developing reliability-aware training objectives and decoding strategies for diffusion-based multimodal generation.


\section*{Acknowledgments}
We acknowledge Advanced Research Computing (ARC) at Virginia Tech for providing the computational resources and technical support that contributed to the results reported in this paper. We thank the reviewers for their constructive feedback, which helped improve this paper.

\bibliography{custom}

@article{li2022diffusion,
  title={Diffusion-LM Improves Controllable Text Generation}, 
  author={Xiang Lisa Li and John Thickstun and Ishaan Gulrajani and Percy Liang and Tatsunori B. Hashimoto},
  journal={Advances in Neural Information Processing Systems},
  volume={35},
  pages={4328--4343},
  year={2022}
}

@article{nie2025large,
  title={Large language diffusion models},
  author={Nie, Shen and Zhu, Fengqi and You, Zebin and Zhang, Xiaolu and Ou, Jingyang and Hu, Jun and Zhou, Jun and Lin, Yankai and Wen, Ji-Rong and Li, Chongxuan},
  journal={Advances in Neural Information Processing Systems},
  volume={38},
  pages={50608--50646},
  year={2026}
}

@inproceedings{liu2025longllada,
  title={Longllada: Unlocking long context capabilities in diffusion llms},
  author={Liu, Xiaoran and Song, Yuerong and Liu, Zhigeng and Huang, Zengfeng and Guo, Qipeng and He, Ziwei and Qiu, Xipeng},
  booktitle={Proceedings of the AAAI Conference on Artificial Intelligence},
  volume={40},
  pages={32186--32194},
  year={2026}
}

@misc{zhu2025llada,
      title={{LLaDA-MoE}: A Sparse MoE Diffusion Language Model}, 
      author={Fengqi Zhu and Zebin You and Yipeng Xing and Zenan Huang and Lin Liu and Yihong Zhuang and Guoshan Lu and Kangyu Wang and Xudong Wang and Lanning Wei and Hongrui Guo and Jiaqi Hu and Wentao Ye and Tieyuan Chen and Chenchen Li and Chengfu Tang and Haibo Feng and Jun Hu and Jun Zhou and Xiaolu Zhang and Zhenzhong Lan and Junbo Zhao and Da Zheng and Chongxuan Li and Jianguo Li and Ji-Rong Wen},
      year={2025},
      eprint={2509.24389},
      archivePrefix={arXiv},
      primaryClass={cs.CL},
      url={https://arxiv.org/abs/2509.24389}, 
}

@misc{ye2025dream,
      title={Dream 7B: Diffusion Large Language Models}, 
      author={Jiacheng Ye and Zhihui Xie and Lin Zheng and Jiahui Gao and Zirui Wu and Xin Jiang and Zhenguo Li and Lingpeng Kong},
      year={2025},
      eprint={2508.15487},
      archivePrefix={arXiv},
      primaryClass={cs.CL},
      url={https://arxiv.org/abs/2508.15487}, 
}

@inproceedings{yang2025mmada,
 author = {Yang, Ling and Tian, Ye and Li, Bowen and Zhang, Xinchen and Shen, Ke and Tong, Yunhai and Wang, Mengdi},
 booktitle = {Advances in Neural Information Processing Systems},
 doi = {10.52202/085713-4636},
 editor = {D. Belgrave and C. Zhang and H. Lin and R. Pascanu and P. Koniusz and M. Ghassemi and N. Chen},
 pages = {138867--138907},
 publisher = {Curran Associates, Inc.},
 title = {MMaDA: Multimodal Large Diffusion Language Models},
 url = {https://proceedings.neurips.cc/paper_files/paper/2025/file/caa934a507a952698d54efb24845fc4b-Paper-Conference.pdf},
 volume = {38, Main Conference},
 year = {2025}
}

@article{li2025lavida,
  title={Lavida: A large diffusion language model for multimodal understanding},
  author={Li, Shufan and Kallidromitis, Konstantinos and Bansal, Hritik and Gokul, Akash and Kato, Yusuke and Kozuka, Kazuki and Kuen, Jason and Lin, Zhe and Chang, Kai-Wei and Grover, Aditya},
  journal={Advances in Neural Information Processing Systems},
  volume={38},
  pages={105101--105134},
  year={2026}
}

@inproceedings{chang2025tracedet,
  title={Tracedet: Hallucination detection from the decoding trace of diffusion large language models},
  author={Chang, Shenxu and Yu, Junchi and Wang, Weixing and Chen, Yongqiang and Yu, Jialin and Torr, Philip and Gu, Jindong},
  booktitle={International Conference on Learning Representations},
  volume={2026},
  pages={9472--9491},
  year={2026}
}

@inproceedings{li2023evaluating,
  title={Evaluating object hallucination in large vision-language models},
  author={Li, Yifan and Du, Yifan and Zhou, Kun and Wang, Jinpeng and Zhao, Wayne Xin and Wen, Ji-Rong},
  booktitle={Proceedings of the 2023 conference on empirical methods in natural language processing},
  pages={292--305},
  year={2023}
}

@inproceedings{karkkainen2021fairface,
  title={Fairface: Face attribute dataset for balanced race, gender, and age for bias measurement and mitigation},
  author={Karkkainen, Kimmo and Joo, Jungseock},
  booktitle={Proceedings of the IEEE/CVF winter conference on applications of computer vision},
  pages={1548--1558},
  year={2021}
}

@inproceedings{atabuzzaman2025benchmarking,
  title={Benchmarking and Mitigating MCQA Selection Bias of Large Vision-Language Models},
  author={Atabuzzaman, Md and Asgarov, Ali and Thomas, Chris},
  booktitle={Proceedings of the 2025 Conference on Empirical Methods in Natural Language Processing},
  pages={33536--33550},
  year={2025}
}

@misc{Qwen2.5-VL,
      title={Qwen2.5-VL Technical Report}, 
      author={Shuai Bai and Keqin Chen and Xuejing Liu and Jialin Wang and Wenbin Ge and Sibo Song and Kai Dang and Peng Wang and Shijie Wang and Jun Tang and Humen Zhong and Yuanzhi Zhu and Mingkun Yang and Zhaohai Li and Jianqiang Wan and Pengfei Wang and Wei Ding and Zheren Fu and Yiheng Xu and Jiabo Ye and Xi Zhang and Tianbao Xie and Zesen Cheng and Hang Zhang and Zhibo Yang and Haiyang Xu and Junyang Lin},
      year={2025},
      eprint={2502.13923},
      archivePrefix={arXiv},
      primaryClass={cs.CV},
      url={https://arxiv.org/abs/2502.13923}, 
}

@misc{internvl2_5,
      title={Expanding Performance Boundaries of Open-Source Multimodal Models with Model, Data, and Test-Time Scaling}, 
      author={Zhe Chen and Weiyun Wang and Yue Cao and Yangzhou Liu and Zhangwei Gao and Erfei Cui and Jinguo Zhu and Shenglong Ye and Hao Tian and Zhaoyang Liu and Lixin Gu and Xuehui Wang and Qingyun Li and Yiming Ren and Zixuan Chen and Jiapeng Luo and Jiahao Wang and Tan Jiang and Bo Wang and Conghui He and Botian Shi and Xingcheng Zhang and Han Lv and Yi Wang and Wenqi Shao and Pei Chu and Zhongying Tu and Tong He and Zhiyong Wu and Huipeng Deng and Jiaye Ge and Kai Chen and Kaipeng Zhang and Limin Wang and Min Dou and Lewei Lu and Xizhou Zhu and Tong Lu and Dahua Lin and Yu Qiao and Jifeng Dai and Wenhai Wang},
      year={2025},
      eprint={2412.05271},
      archivePrefix={arXiv},
      primaryClass={cs.CV},
      url={https://arxiv.org/abs/2412.05271}, 
}

@article{llava,
  title={Visual instruction tuning},
  author={Liu, Haotian and Li, Chunyuan and Wu, Qingyang and Lee, Yong Jae},
  journal={Advances in Neural Information Processing Systems},
  volume={36},
  year={2023}
}

@article{instructblip,
      title={InstructBLIP: Towards General-purpose Vision-Language Models with Instruction Tuning}, 
      author={Wenliang Dai and Junnan Li and Dongxu Li and Anthony Meng Huat Tiong and Junqi Zhao and Weisheng Wang and Boyang Li and Pascale Fung and Steven Hoi},
      journal={Advances in Neural Information Processing Systems},
      volume = {37},
      year={2023}
}

@article{gpt4o,
      title={GPT-4o System Card}, 
      author={OpenAI},
      journal={arXiv preprint arXiv:2410.21276},
      year={2024}
}

@techreport{wah2011caltech,
	Title = {The Caltech-UCSD Birds-200-2011 Dataset},
	Author = {Wah, C. and Branson, S. and Welinder, P. and Perona, P. and Belongie, S.},
	Year = {2011},
	Institution = {California Institute of Technology},
	Number = {CNS-TR-2011-001}
}

@inproceedings{khosla2011novel,
  title={Novel dataset for fine-grained image categorization: Stanford dogs},
  author={Khosla, Aditya and Jayadevaprakash, Nityananda and Yao, Bangpeng and Li, Fei-Fei},
  booktitle={Proc. CVPR workshop on fine-grained visual categorization (FGVC)},
  volume={2},
  year={2011}
}

@inproceedings{pezeshkpour2024large,
  title={Large Language Models Sensitivity to The Order of Options in Multiple-Choice Questions},
  author={Pezeshkpour, Pouya and Hruschka, Estevam},
  booktitle={Findings of the Association for Computational Linguistics: NAACL 2024},
  pages={2006--2017},
  year={2024}
}

@article{zheng2023judging,
  title={Judging llm-as-a-judge with mt-bench and chatbot arena},
  author={Zheng, Lianmin and Chiang, Wei-Lin and Sheng, Ying and Zhuang, Siyuan and Wu, Zhanghao and Zhuang, Yonghao and Lin, Zi and Li, Zhuohan and Li, Dacheng and Xing, Eric and others},
  journal={Advances in Neural Information Processing Systems},
  volume={36},
  pages={46595--46623},
  year={2023}
}

@inproceedings{wang2021gender,
  title={Are Gender-Neutral Queries Really Gender-Neutral? Mitigating Gender Bias in Image Search},
  author={Wang, Jialu and Liu, Yang and Wang, Xin},
  booktitle={Proceedings of the 2021 Conference on Empirical Methods in Natural Language Processing},
  pages={1995--2008},
  year={2021}
}

@inproceedings{zhao2025large,
  title={Large language models badly generalize across option length, problem types, and irrelevant noun replacements},
  author={Zhao, Guangxiang and Hu, Saier and Jian, Xiaoqi and Jinzhu, Wu and Wu, Yuhan and Sun, Lin and Zhang, Xiangzheng},
  booktitle={Proceedings of the 2025 Conference on Empirical Methods in Natural Language Processing},
  pages={26825--26834},
  year={2025}
}

@inproceedings{you2025llada,
  title={Llada-v: Large language diffusion models with visual instruction tuning},
  author={You, Zebin and Nie, Shen and Zhang, Xiaolu and ZHOU, JUN and Lu, Zhiwu and Wen, Ji-Rong and Li, Chongxuan},
  booktitle={Proceedings of the IEEE/CVF Conference on Computer Vision and Pattern Recognition},
  pages={10093--10105},
  year={2026}
}

@inproceedings{bang2023multitask,
  title={A multitask, multilingual, multimodal evaluation of chatgpt on reasoning, hallucination, and interactivity},
  author={Bang, Yejin and Cahyawijaya, Samuel and Lee, Nayeon and Dai, Wenliang and Su, Dan and Wilie, Bryan and Lovenia, Holy and Ji, Ziwei and Yu, Tiezheng and Chung, Willy and others},
  booktitle={Proceedings of the 13th international joint conference on natural language processing and the 3rd conference of the asia-pacific chapter of the association for computational linguistics (volume 1: Long papers)},
  pages={675--718},
  year={2023}
}

@article{ye2023mplug,
  title={{mPLUG-Owl}: Modularization Empowers Large Language Models with Multimodality}, 
      author={Qinghao Ye and Haiyang Xu and Guohai Xu and Jiabo Ye and Ming Yan and Yiyang Zhou and Junyang Wang and Anwen Hu and Pengcheng Shi and Yaya Shi and Chenliang Li and Yuanhong Xu and Hehong Chen and Junfeng Tian and Qi Qian and Ji Zhang and Fei Huang and Jingren Zhou},
  journal={arXiv preprint arXiv:2304.14178},
  year={2023}
}

@inproceedings{zhu2024minigpt,
  title={MiniGPT-4: Enhancing Vision-Language Understanding with Advanced Large Language Models}, 
  author={Deyao Zhu and Jun Chen and Xiaoqian Shen and Xiang Li and Mohamed Elhoseiny},
  booktitle={International Conference on Learning Representations (ICLR)},
  volume={2024},
  pages={18378--18394},
  year={2024}
}

@inproceedings{berglund2024reversal,
  title={The Reversal Curse: {LLM}s trained on “{A} is {B}” fail to learn “{B} is {A}”},
  author={Berglund, Lukas and Tong, Meg and Kaufmann, Maximilian and Balesni, Mikita and Stickland, Asa and Korbak, Tomek and Evans, Owain},
  booktitle={International Conference on Learning Representations},
  volume={2024},
  pages={18623--18642},
  year={2024}
}

@misc{yu2025dimple,
      title={Dimple: Discrete Diffusion Multimodal Large Language Model with Parallel Decoding}, 
      author={Runpeng Yu and Xinyin Ma and Xinchao Wang},
      year={2025},
      eprint={2505.16990},
      archivePrefix={arXiv},
      primaryClass={cs.CV},
      url={https://arxiv.org/abs/2505.16990}, 
}

@article{ye2025dreamV,
  title={{Dream-VL} \& {Dream-VLA}: Open Vision-Language and Vision-Language-Action Models with Diffusion Language Model Backbone},
  author={Ye, Jiacheng and Gong, Shansan and Gao, Jiahui and Fan, Junming and Wu, Shuang and Bi, Wei and Bai, Haoli and Shang, Lifeng and Kong, Lingpeng},
  journal={arXiv preprint arXiv:2512.22615},
  year={2025}
}

@inproceedings{rohrbach2018object,
  title={Object hallucination in image captioning},
  author={Rohrbach, Anna and Hendricks, Lisa Anne and Burns, Kaylee and Darrell, Trevor and Saenko, Kate},
  booktitle={Proceedings of the 2018 Conference on Empirical Methods in Natural Language Processing},
  pages={4035--4045},
  year={2018}
}

@inproceedings{liu2024improved,
  title={Improved baselines with visual instruction tuning},
  author={Liu, Haotian and Li, Chunyuan and Li, Yuheng and Lee, Yong Jae},
  booktitle={Proceedings of the IEEE/CVF conference on computer vision and pattern recognition},
  pages={26296--26306},
  year={2024}
}

@inproceedings{lin2014microsoft,
  title={Microsoft coco: Common objects in context},
  author={Lin, Tsung-Yi and Maire, Michael and Belongie, Serge and Hays, James and Perona, Pietro and Ramanan, Deva and Doll{\'a}r, Piotr and Zitnick, C Lawrence},
  booktitle={European conference on computer vision},
  pages={740--755},
  year={2014},
  organization={Springer}
}

\appendix

\section{Appendix}
\label{sec:appendix}

This section contains the following topics in detail. 
\begin{itemize}
    \item Additional Qualitative Hallucination Analysis (Appendix \ref{sec:qualitative_app})
    \item Attention-Based Analysis (Appendix \ref{sec:attention_analysis})
    \item Demographic Bias Evaluation Protocol (Appendix \ref{sec:fairface_protocol})
    \item Demographic Bias Evaluation Results (Appendix \ref{sec:fairface_results})
    \item Demographic Bias Confusion Summary (Appendix \ref{sec:demographic_confusion})
    \item Selection Bias Comparison (Appendix \ref{sec:selection_comp})
    \item Linguistic Quality Evaluation Prompt (Appendix \ref{sec:linguistic_prompt})
\end{itemize}



\begin{figure*}[!ht]
    \centering
    \includegraphics[width=\linewidth]{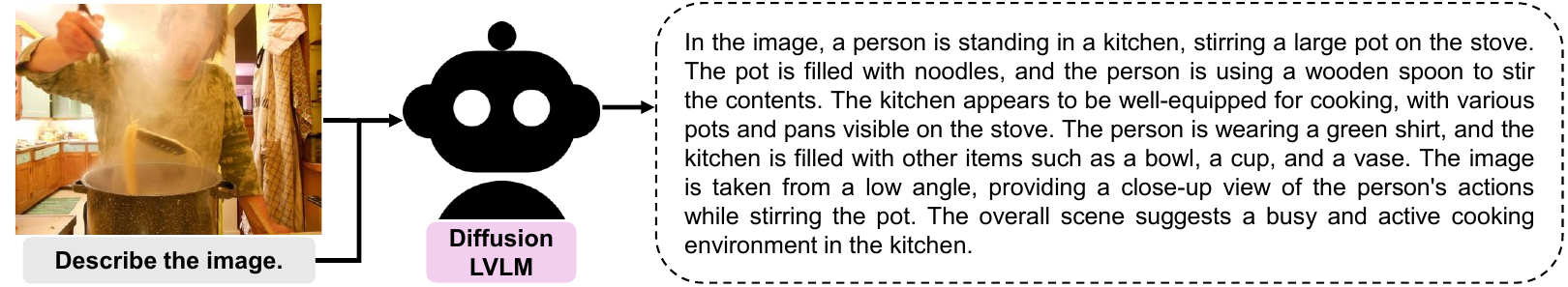}
    \vspace{0.3em}
    \includegraphics[width=\linewidth]{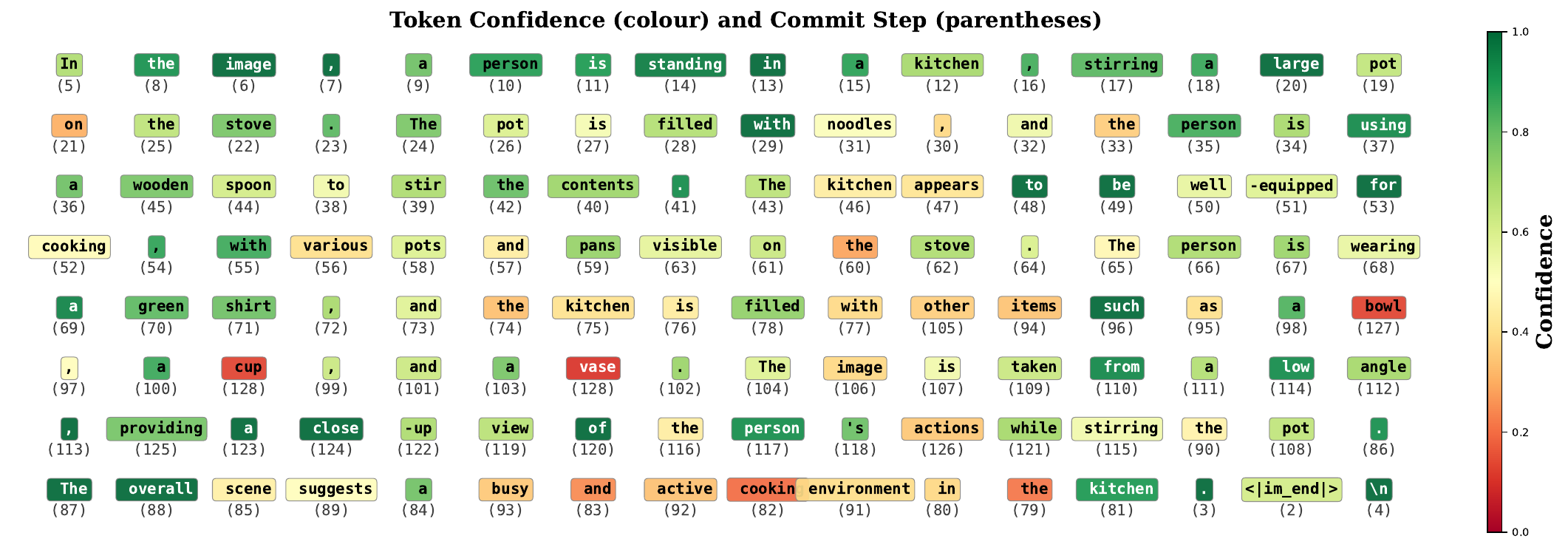}
    \vspace{-2.5em}
    \caption{Qualitative hallucination analysis on a LaViDa-Dream caption (MSCOCO). \textbf{Top:} The model generates a fluent description of a kitchen cooking scene, but hallucinates objects absent from the image (\textit{bowl}, \textit{cup}, \textit{vase}). \textbf{Bottom:} Per-token confidence (colour) and commit step (parentheses). All three hallucinated tokens are committed at the latest denoising steps with the lowest confidence in the sequence: \textit{bowl} (step~127), \textit{cup} (step~128), and \textit{vase} (step~128), all shown in red. In contrast, accurately grounded tokens such as \textit{person} (step~10), \textit{kitchen} (step~12), \textit{stirring} (step~17), and \textit{noodles} (step~31) are committed early with high confidence (dark green). The clustering of all three hallucinated objects at steps 127--128 with near-zero confidence, while visually accurate content words are committed in the first 35 steps, provides strong evidence that late commit step combined with low confidence jointly indicate hallucination risk in diffusion-based generation.}
    \label{fig:qualitative_hallucination_lavida_dream}
\end{figure*}

\begin{figure*}[!ht]
    \centering
    \includegraphics[width=\linewidth]{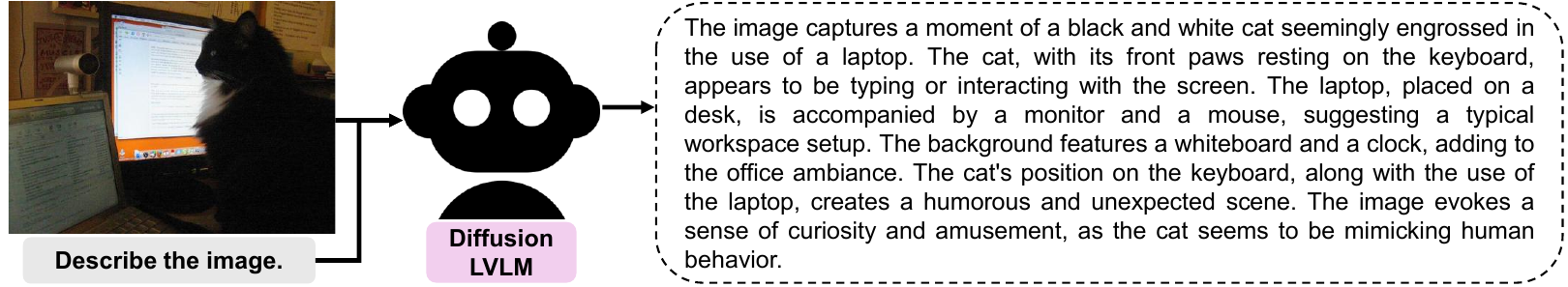}
    \vspace{0.3em}
    \includegraphics[width=\linewidth]{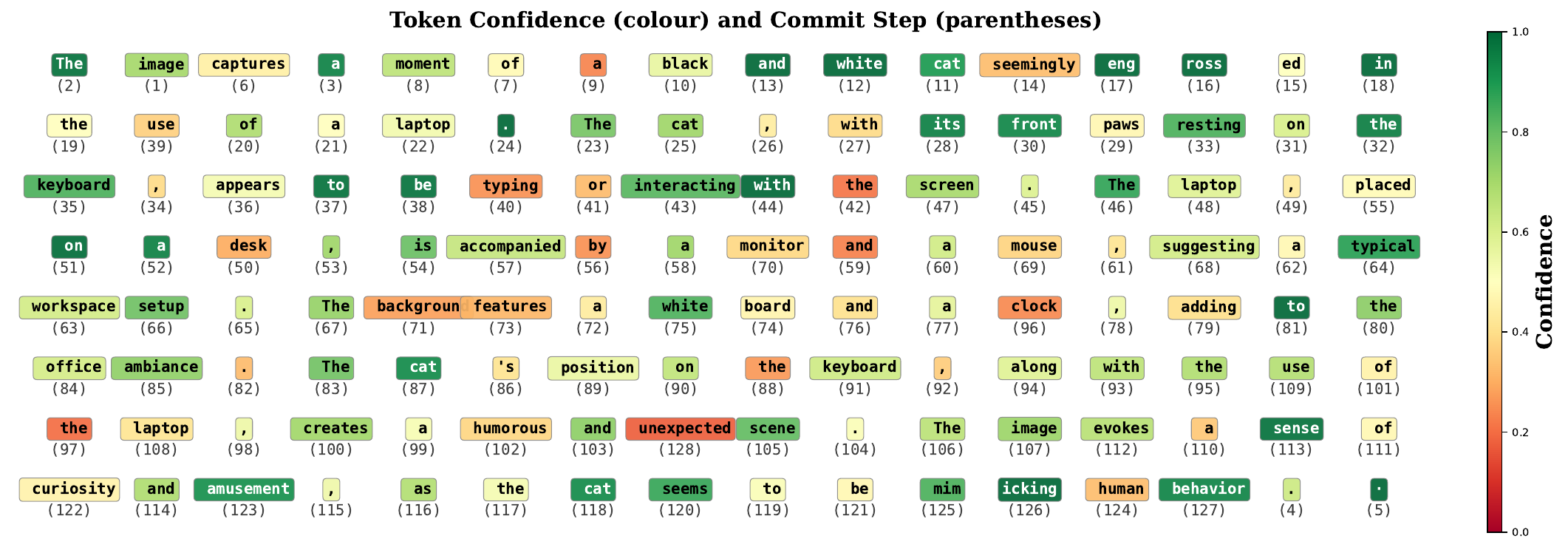}
    \vspace{-2.5em}
    \caption{Qualitative hallucination analysis on a LaViDa-LLaDA caption (MSCOCO). \textbf{Top:} The model generates a fluent description of a cat-and-laptop scene, but hallucinates objects absent from the image (\textit{clock},
    \textit{whiteboard}). \textbf{Bottom:} Per-token confidence (colour) and commit step (parentheses). Hallucinated tokens \textit{clock} (step~96, low confidence) and \textit{unexpected} (step~128, lowest confidence in the sequence) are committed at late denoising steps, whereas accurately grounded tokens such as \textit{cat} (step~25), \textit{laptop} (step~22), and \textit{keyboard} (step~35) are committed early with high confidence. This pattern is consistent with the hypothesis that late commit step combined with low confidence jointly indicate hallucination risk in dLVLMs.}
    \label{fig:qualitative_hallucination_app}
\end{figure*}


\subsection{Additional Qualitative Hallucination Analysis} \label{sec:qualitative_app} 

Figure~\ref{fig:qualitative_hallucination_lavida_dream} presents a second qualitative example using LaViDa-Dream on a kitchen cooking scene, further examining the commit-step hallucination hypothesis introduced in Section~\ref{sec:hallucination}. The model correctly grounds early-committed tokens such as \textit{person} (step~10), \textit{kitchen} (step~12), \textit{stirring} (step~17), and \textit{noodles} (step~31) with high confidence, all consistent with the visual content. In contrast, all three hallucinated objects --- \textit{bowl} (step~127), \textit{cup} (step~128), and \textit{vase} (step~128) --- are committed at the final denoising steps with near-zero confidence (red), despite none of these objects being present in the image. The clustering of all three hallucinations at steps 127--128 is particularly striking, suggesting that the model defaults to generic kitchen objects under maximum uncertainty at the end of the denoising process. 

Figure~\ref{fig:qualitative_hallucination_app} presents a third example on a distinct scene type (an indoor desk with a cat), where \textit{clock} (step~96, low confidence) is hallucinated while accurately grounded tokens such as \textit{cat} (step~25), \textit{laptop} (step~22), and \textit{keyboard} (step~35) are committed early with high confidence. Together, these three examples across distinct image types, scene categories, and diffusion backbones (LaViDa-LLaDA and LaViDa-Dream) provide convergent evidence that late commit step combined with low confidence may generalize as a proxy signal for hallucination risk in diffusion-based generation.


\subsection{Attention-Based Analysis}
\label{sec:attention_analysis}
To further investigate the grounding failure hypothesis, we examined whether hallucinated tokens attend differently to image patch tokens compared to grounded tokens in LaViDa-LLaDA. We registered forward hooks on query and key projections across three representative layers (8, 16, 24) and extracted attention weights from answer token positions to the 718 image patch positions, running inference without prefix caching to ensure answer tokens attend over the full sequence at every denoising step. We examined attention both averaged across all denoising steps and targeted specifically at each token's commit step, computing mean attention, peak attention, and entropy over the image region for hallucinated versus grounded tokens. In both settings, we found no meaningful difference between hallucinated and grounded tokens across all three layers: mean attention values were nearly identical (0.0005 vs.\ 0.0005), entropy showed no consistent pattern (4.41 vs.\ 4.40), and peak attention was similarly uninformative. Notably, however, the commit steps themselves remain highly discriminative in this setting as well: grounded tokens such as \textit{sofa} are committed at step 19 (Figure~\ref{fig:qualitative_hallucination}), while hallucinated tokens such as \textit{clock} and \textit{vase} are committed at steps 127 and 126 respectively, consistent with the commit-step hypothesis introduced above. This suggests that the commit step and confidence trajectory remain the primary mechanistic signals for hallucination in dLVLMs, while attention patterns over image tokens do not provide additional discriminative power at the individual token level. A more comprehensive analysis incorporating spatial attention map visualization, head-level examination, and evaluation across multiple models remains an important direction for future work.


\subsection{Demographic Bias Evaluation Protocol}
\label{sec:fairface_protocol}

Gender classification used the completion-style prompt: \textit{``Look at the person in this image. Then tell what is the gender of the person. Respond with 
a single word only, either `male' or `female'. Do not write any other words, 
punctuation, or explanation. Answer:''} Race recognition was framed as a 
multiple-choice question presenting all seven FairFace labels (East Asian, 
Indian, Black, White, Middle Eastern, Latino Hispanic, Southeast Asian) as 
lettered options (A--G), with the instruction to respond with only the letter 
of the best answer. For gender, outputs were exact-matched against 
\textit{male}/\textit{female} after lowercasing; for race, the selected letter 
was mapped back to the corresponding FairFace label. Non-matching outputs 
were counted as incorrect, making the evaluation conservative. Both mappings 
are deterministic with no ambiguity.


\subsection{Demographic Bias Evaluation Results}
\label{sec:fairface_results}

Table~\ref{tab:fairface_bias} presents the full gender and race recognition results on FairFace under two face-crop padding settings. Results are reported for all evaluated models across seven racial groups and two genders, with the female--male accuracy gap and per-model accuracy change across padding settings ($\Delta$). Bold entries indicate the best result per column; underlined entries indicate the second best; highlighted entries indicate the best among diffusion models. $\Delta$ rows are shaded in gray to distinguish them from primary results. A detailed analysis of these results is provided in Section~\ref{sec:demo_bias_main}.

\begin{table*}[!ht]
\centering
\resizebox{\textwidth}{!}{%
\begin{tabular}{l l c
  ccc c
  |
  ccccccc c
}
\toprule
\multirow{2}{*}{\textbf{Type}} &
\multirow{2}{*}{\textbf{Model}} &
\multirow{2}{*}{\textbf{Pad.}} &
\multicolumn{4}{c|}{\textbf{Gender Accuracy (\%)}} &
\multicolumn{8}{c}{\textbf{Per-Race Accuracy (\%)}} \\
\cmidrule(lr){4-7} \cmidrule(lr){8-15}
& & &
\textbf{Female} & \textbf{Male} & \textbf{Overall} & \textbf{Gap} &
\textbf{Black} & \textbf{E.Asian} & \textbf{Indian} &
\textbf{Latino} & \textbf{M.East} & \textbf{SE.Asian} & \textbf{White} &
\textbf{Overall} \\
\midrule

\multirow{9}{*}{AR}
  & \multirow{2}{*}{LLaVA}
  & 0.25
  & 94.77 & \underline{97.48} & 96.13 & $-$2.71
  & 88.73 & \textbf{98.94} & 73.24 & 24.30 & 22.18 & 8.45 & 79.58 & 56.49 \\
  & & 1.25
  & 94.47 & 97.28 & 95.88 & $-$2.81
  & 83.80 & 96.13 & \underline{74.65} & 51.76 & 33.10 & 19.72 & 61.97 & 59.61 \\ 
\rowcolor{gray!20}
\cellcolor{white} &\cellcolor{white} & $\Delta$ & \textit{$-$0.30} & \textit{$-$0.20} & \textit{$-$0.25} & --
  & \textit{$-$4.93} & \textit{$-$2.81} & \textit{$+$1.41} & \textit{$+$27.46} & \textit{$+$10.92} & \textit{$+$11.27} & \textit{$-$17.61} & \textit{$+$3.12} \\
\addlinespace[2pt]

& \multirow{2}{*}{Qwen2.5-VL} 
  & 0.25
  & 94.67 & 91.45 & 93.06 & $+$3.22
  & 88.03 & 88.73 & 71.48 & 21.13 & 48.94 & 35.56 & \textbf{86.62} & 62.93 \\
  & & 1.25
  & \underline{98.39} & 95.37 & \underline{96.88} & $+$3.02
  & 86.97 & 87.32 & \textbf{76.06} & 39.08 & \underline{54.23} & \underline{51.06} & 84.15 & \underline{68.41} \\
\rowcolor{gray!20}
\cellcolor{white} &\cellcolor{white} & $\Delta$ & \textit{$+$3.72} & \textit{$+$3.92} & \textit{$+$3.82} & --
  & \textit{$-$1.06} & \textit{$-$1.41} & \textit{$+$4.58} & \textit{$+$17.95} & \textit{$+$5.29} & \textit{$+$15.50} & \textit{$-$2.47} & \textit{$+$5.48} \\
\addlinespace[2pt]

& \multirow{2}{*}{InternVL2.5}
  & 0.25
  & 93.06 & 94.06 & 93.56 & $-$1.00
  & \underline{92.61} & 90.85 & 52.46 & 50.00 & 45.07 & 19.37 & 82.04 & 61.77 \\
  & & 1.25
  & 95.88 & \textbf{97.59} & 96.73 & $-$1.71
  & 90.14 & 91.90 & 70.42 & \textbf{53.52} & \textbf{57.75} & 41.55 & 79.93 & \textbf{69.32} \\
\rowcolor{gray!20}
\cellcolor{white} &\cellcolor{white} & $\Delta$ & \textit{$+$2.82} & \textit{$+$3.53} & \textit{$+$3.17} & --
  & \textit{$-$2.47} & \textit{$+$1.05} & \textit{$+$17.96} & \textit{$+$3.52} & \textit{$+$12.68} & \textit{$+$22.18} & \textit{$-$2.11} & \textit{$+$7.55} \\ 
\midrule

\multirow{16}{*}{Diff.}
  & \multirow{2}{*}{LLaDA-V}
  & 0.25
  & 73.14 & 93.96  & 83.55  & $-$ 20.82
  & 61.97  & 44.72  & 24.30  & 5.28  & 16.90  & 39.79  & 61.97  & 36.42 \\
  & & 1.25
  & 96.98  & 94.47  & 95.72 & $+$2.51
  &  54.23  & 80.99  & 35.56  & 23.24  & 28.87  & \cellcolor{diffbest}52.46  & 57.04  & 47.48 \\
\rowcolor{gray!20}
\cellcolor{white} &\cellcolor{white} &$\Delta$ & \textit{$+$23.84} & \textit{$+$0.51} & \textit{$+$12.17} & --
  & \textit{$-$7.74} & \textit{$+$36.27} & \textit{$+$11.26} & \textit{$+$17.96} & \textit{$+$11.97} & \textit{$+$12.67} & \textit{$-$4.93} & \textit{$+$11.06} \\
\addlinespace[2pt]

  & \multirow{2}{*}{LaViDa-D}
  & 0.25
  & 96.68 & 83.20 & 89.94 & $+$13.48
  & 85.56 & \cellcolor{diffbest}\underline{97.54} & 25.35 & 1.06 & 16.90 & 4.58 & 55.63 & 40.95 \\
  & & 1.25
  & 97.99 & 92.15 & 95.07 & $+$5.84
  & \cellcolor{diffbest}\textbf{95.07} & 96.13 & 49.30 & 11.62 & 23.94 & 5.28 & 61.97 & 49.04 \\
\rowcolor{gray!20}
\cellcolor{white} &\cellcolor{white} & $\Delta$ & \textit{$+$1.31} & \textit{$+$8.95} & \textit{$+$5.13} & --
  & \textit{$+$9.51} & \textit{$-$1.41} & \textit{$+$23.95} & \textit{$+$10.56} & \textit{$+$7.04} & \textit{$+$0.70} & \textit{$+$6.34} & \textit{$+$8.09} \\
\addlinespace[2pt]

& \multirow{2}{*}{LaViDa-L}
  & 0.25
  & 90.44 & 80.18 & 85.31 & $+$10.26
  & 68.31 & 76.76 & 64.08 & 0.35 & 8.80 & 1.06 & 64.44 & 40.54 \\
  & & 1.25
  & 97.59 & 93.86 & 95.72 & $+$3.73
  & 73.24 & 80.99 & 72.89 & 5.99 & 20.07 & 23.24 & 78.17 & 50.65 \\
\rowcolor{gray!20}
\cellcolor{white} &\cellcolor{white} & $\Delta$
  & \textit{$+$7.15} & \textit{$+$13.68} & \textit{$+$10.41} & --
  & \textit{$+$4.93} & \textit{$+$4.23} & \textit{$+$8.81} & \textit{$+$5.64} & \textit{$+$11.27} & \textit{$+$22.18} & \textit{$+$13.73} & \textit{$+$10.11} \\
\addlinespace[2pt]

  & \multirow{2}{*}{MMaDA-M}
  & 0.25
  & 68.51 & 91.85 & 80.18 & $-$23.34
  & 46.48 & 7.39 & 34.51 & 0.00 & 24.65 & 0.00 & 48.59 & 23.09 \\
  & & 1.25
  & 93.96 & 93.86 & 93.91 & $+$0.10
  & 52.11 & 31.69 & 24.30 & 2.46 & 21.13 & 0.00 & 35.21 & 23.84 \\
\rowcolor{gray!20}
\cellcolor{white} &\cellcolor{white} & $\Delta$ 
 & \textit{$+$25.45} & \textit{$+$2.01} & \textit{$+$13.73} & --
  & \textit{$+$5.63} & \textit{$+$24.30} & \textit{$-$10.21} & \textit{$+$2.46} & \textit{$-$3.52} & \textit{0.00} & \textit{$-$13.38} & \textit{$+$0.75} \\
\addlinespace[2pt]

  & \multirow{2}{*}{Dream-VL}
  & 0.25
  & 92.76  & 95.37  & 94.06 & $-$2.61
  & 89.44  & 90.49 & 67.61  & 42.96  & 44.72  & 31.69 & 79.23 & 63.73  \\
  & & 1.25
  & 97.28  & \cellcolor{diffbest}97.08  & \cellcolor{diffbest}\textbf{97.18}  & $+$0.20
  & 91.90 & 93.66 & \cellcolor{diffbest}\textbf{76.06}  & \cellcolor{diffbest}\underline{52.46}  & \cellcolor{diffbest}47.89  & 35.92  & 78.87  & \cellcolor{diffbest}68.11  \\
\rowcolor{gray!20}
\cellcolor{white} &\cellcolor{white} & $\Delta$
  & \textit{$+$4.52} & \textit{$+$1.71} & \textit{$+$3.12} & --
  & \textit{$+$2.46} & \textit{$+$3.17} & \textit{$+$8.45} & \textit{$+$9.50} & \textit{$+$3.17} & \textit{$+$4.23} & \textit{$-$0.36} & \textit{$+$4.38} \\
  \addlinespace[2pt]

  & \multirow{2}{*}{Dimple}
  & 0.25
  & 95.67  & 88.73  & 92.20 & $+$6.94
  & 88.73  & 90.14 & 36.97  & 3.17  & 16.90  & 11.27  & \cellcolor{diffbest}\underline{86.27}  & 47.64  \\
  & & 1.25
  & \cellcolor{diffbest}\textbf{98.49}  & 95.17  & 96.83  & $+$3.32
  & 92.25 & 92.61  & 55.63  & 25.00  & 38.38  & 33.80  & 79.23  & 59.56  \\
\rowcolor{gray!20}
\cellcolor{white} &\cellcolor{white} & $\Delta$
  & \textit{$+$2.82} & \textit{$+$6.44} & \textit{$+$4.63} & --
  & \textit{$+$3.52} & \textit{$+$2.47} & \textit{$+$18.66} & \textit{$+$21.83} & \textit{$+$21.48} & \textit{$+$22.53} & \textit{$-$7.04} & \textit{$+$11.92} \\

\bottomrule
\end{tabular}%
}
\vspace{-0.75em}
\caption{
  Gender and race recognition accuracy (\%) on FairFace under two face-crop padding settings (0.25 = tight crop; 1.25 = loose crop with background context). \textbf{Bold} = best per column; \underline{underline} = second best; \colorbox{diffbest}{\strut highlighted} = best among diffusion models. Gap = Female $-$ Male accuracy (positive = female-favoring; negative = male-favoring). $\Delta$ = accuracy change from padding $0.25 \to 1.25$; $\Delta$ rows are shaded in gray. E.Asian = East Asian; M.East = Middle Eastern; SE.Asian = Southeast Asian; Latino = Latino Hispanic.
}
\label{tab:fairface_bias}
\end{table*}


\begin{table*}[!ht]
\centering
\resizebox{\textwidth}{!}{%
\begin{tabular}{l ll ll ll ll}
\toprule
 &
\multicolumn{2}{c}{\textbf{LLaVA-1.5-7B}} &
\multicolumn{2}{c}{\textbf{LaViDa-Dream}} &
\multicolumn{2}{c}{\textbf{LaViDa-LLaDA}} &
\multicolumn{2}{c}{\textbf{MMaDA-MixCoT}} \\
\cmidrule(lr){2-3} \cmidrule(lr){4-5} \cmidrule(lr){6-7} \cmidrule(lr){8-9}
\textbf{True Race} &
\textbf{\#1} & \textbf{\#2} &
\textbf{\#1} & \textbf{\#2} &
\textbf{\#1} & \textbf{\#2} &
\textbf{\#1} & \textbf{\#2} \\
\midrule

Black
  & \textbf{Blk(89\%)} & Wht(3\%)
  & \textbf{Blk(86\%)} & EAs(11\%)
  & \textbf{Blk(68\%)} & Ind(27\%)
  & \textbf{Blk(46\%)} & Ind(31\%) \\

East Asian
  & \textbf{EAs(99\%)} & Ind(0\%)
  & \textbf{EAs(98\%)} & Blk(2\%)
  & \textbf{EAs(77\%)} & Wht(9\%)
  & Wht(32\%)           & Ind(25\%) \\

Indian
  & \textbf{Ind(73\%)} & EAs(9\%)
  & EAs(40\%)           & \textbf{Ind(25\%)}
  & \textbf{Ind(64\%)} & Blk(18\%)
  & \textbf{Ind(35\%)} & Blk(24\%) \\

Latino
  & Wht(37\%)           & \textbf{Lat(24\%)}
  & EAs(63\%)           & Blk(19\%)
  & Ind(37\%)           & Wht(24\%)
  & Ind(35\%)           & Wht(26\%) \\

Middle Eastern
  & Wht(45\%)           & \textbf{MEa(22\%)}
  & EAs(44\%)           & Wht(26\%)
  & Wht(35\%)           & Ind(25\%)
  & Wht(38\%)           & \textbf{MEa(25\%)} \\

SE Asian
  & EAs(73\%)           & MEa(14\%)
  & EAs(84\%)           & MEa(6\%)
  & EAs(42\%)           & Ind(32\%)
  & Wht(33\%)           & Ind(30\%) \\

White
  & \textbf{Wht(80\%)} & EAs(13\%)
  & \textbf{Wht(56\%)} & EAs(34\%)
  & \textbf{Wht(64\%)} & EAs(18\%)
  & \textbf{Wht(49\%)} & Ind(18\%) \\

\bottomrule
\end{tabular}%
}
\caption{
  Prediction distribution (top-2 predicted races) for each true racial group
  on FairFace (padding = 0.25). Correct predictions are \textbf{bolded}.
  Numbers in parentheses are the fraction of predictions in that category. \textbf{Abbreviations:} Blk = Black; EAs = East Asian; Ind = Indian; Lat = Latino Hispanic; MEa = Middle Eastern; SEA = Southeast Asian; Wht = White. All results use padding 0.25 (tight face crop).
}
\label{tab:race_confusion}
\end{table*}


\subsection{Demographic Bias Confusion Summary}
\label{sec:demographic_confusion}

Table~\ref{tab:race_confusion} reveals systematic misclassification patterns that are consistent with training data bias rather than random error. Across all models, Latino Hispanic and Southeast Asian faces are the most severely misclassified: LaViDa-Dream absorbs 63\% of Latino Hispanic faces into East Asian and 84\% of Southeast Asian faces into East Asian, while LLaVA-1.5-7B redirects 73\% of Southeast Asian faces to East Asian. These absorption patterns indicate that models conflate visually proximate racial groups into dominant prototypes, a failure mode shared across both AR and diffusion backbones. MMaDA-MixCoT exhibits a distinct pattern, defaulting heavily to White and Indian across nearly all misclassified groups, consistent with its near-zero accuracy on East Asian (7.39\%) and Latino Hispanic (0.00\%) at tight crop.

Backbone-specific confusion patterns further implicate the generative mechanism in shaping racial representations. LaViDa-Dream and LaViDa-LLaDA share the same training pipeline yet show qualitatively different top-2 confusion targets: LaViDa-Dream defaults to East Asian as its dominant misclassification category across Indian (40\%), Latino Hispanic (63\%), Middle Eastern (44\%), and Southeast Asian (84\%) groups, whereas LaViDa-LLaDA instead defaults to Indian and White, achieving substantially higher Indian accuracy (64.08\% vs.\ 25.35\%) but lower East Asian accuracy (76.76\% vs.\ 97.54\%). This divergence, arising purely from the choice of diffusion backbone (Dream-7B vs.\ LLaDA-8B), suggests that the generative backbone encodes distinct racial prototype hierarchies independent of the shared vision encoder and fine-tuning data.


\subsection{Selection Bias Comparison}
\label{sec:selection_comp}


\begin{table}[t]
\centering
\begingroup
\setlength{\tabcolsep}{3.2pt}
\renewcommand{\arraystretch}{0.95}
\scriptsize
\resizebox{\columnwidth}{!}{%

\begin{tabular}{ll|cccc}
\toprule
\textbf{\makecell{Model\\Type}} & \textbf{Model} &
\textbf{\makecell{Equal\\Long}} &
\textbf{\makecell{Equal\\Short}} &
\textbf{\makecell{Shorter\\Correct}} &
\textbf{\makecell{Longer\\Correct}} \\
\midrule

\multicolumn{6}{l}{\textbf{CUB With Class Name}} \\
\cmidrule(lr){1-6}

\multirow{4}{*}{AR} & LLaVA-v1.5-7B  & 27.50 & 28.70 & 23.40 & 63.50 \\
 & LLaVA-v1.5-13B  &28.30 & 33.70 & 19.10 &76.40 \\
 & Qwen2.5-VL-7B  & \textbf{76.10} &\textbf{ 74.30} & \textbf{50.70} & 93.90 \\
 & InternVL-2.5-8B  & 61.40 & 60.00 & 39.20 & 88.30 \\ 
 \cmidrule(lr){1-6}
\multirow{6}{*}{Diff.} & LLaDA-V       & 60.70 & 60.10 & 37.40 & 92.50 \\
 & LaViDa-LLaDA  & 47.40 & 48.60 & 11.50 & \textbf{96.60} \\
 & LaViDa-Dream  & 45.10 & 46.50 & 9.00  & 92.60 \\
 & MMaDA-MixCoT  & 37.40 & 40.80 & 11.40 & 91.70 \\
 & Dream-VL     & 68.20  & 68.40  & 42.10  & 93.00 \\
 & Dimple      & 54.30  & 54.70  & 26.80  & 87.10 \\

\cmidrule(lr){1-6}
\multicolumn{6}{l}{\textbf{CUB Without Class Name}} \\
\cmidrule(lr){1-6}

\multirow{4}{*}{AR} & LLaVA-v1.5-7B   & 26.70 & 31.20 & 8.90 & 96.80 \\
 & LLaVA-v1.5-13B  & 39.30 & 42.60 & 15.20 & 93.30 \\
 & Qwen2.5-VL-7B   & \textbf{64.90} & \textbf{61.00} & 26.60 & 93.40 \\
 & InternVL-2.5-8B & 62.50 & 57.10 & 37.80 & 88.30 \\ \cmidrule(lr){1-6}
\multirow{6}{*}{Diff.} & LLaDA-V       & 63.60 & 51.30 & \textbf{47.00} & 82.30 \\
 & LaViDa-LLaDA  & 51.90 & 48.10 & 6.40 & \textbf{97.70} \\
 & LaViDa-Dream  & 38.60 & 48.90 & 2.80 & 96.50 \\
 & MMaDA-MixCoT   & 39.00 & 42.10 & 12.80 & 89.70 \\
 & Dream-VL   & 63.00  & 57.90  & 30.70   & 84.40  \\ 
 & Dimple  & 50.00  & 49.90  & 19.70  & 80.90  \\

\cmidrule(lr){1-6}
\multicolumn{6}{l}{\textbf{Dog With Class Name}} \\
\cmidrule(lr){1-6}

\multirow{4}{*}{AR} & LLaVA-v1.5-7B       & 26.67 & 31.83 & 16.00 & 76.67 \\
 & LLaVA-v1.5-13B      & 30.83 & 37.67 & 20.00 & 76.83 \\
 & Qwen2.5-VL-7B       & \textbf{76.33} & \textbf{78.00} & \textbf{58.83} & \textbf{93.33} \\
 & InternVL-2.5-8B     & 59.33 & 57.67 & 43.67 & 79.33 \\ \cmidrule(lr){1-6}

\multirow{5}{*}{Diff.} & LLaDA-V      & 57.83 & 55.33 & 45.50 & 82.67 \\
 & LaViDa-LLaDA & 48.50 & 44.33 & 21.67 & 92.67 \\
 & LaViDa-Dream & 59.00 & 52.50 & 21.50 & 91.67 \\
 & MMaDA-MixCoT & 37.33 & 37.33 & 24.50 & 72.17 \\
 & Dream-VL     & 72.33 & 66.50 & 56.00  & 87.33  \\

\cmidrule(lr){1-6}
\multicolumn{6}{l}{\textbf{Dog Without Class Name}} \\
\cmidrule(lr){1-6}
\addlinespace[2pt]

\multirow{4}{*}{AR} & LLaVA-v1.5-7B        & 30.00 & 33.67 & 5.67 & \textbf{99.17} \\
 & LLaVA-v1.5-13B       & 36.67 & 42.33 & 14.00 & 95.67 \\
 & Qwen2.5-VL-7B        & \textbf{62.83} & 58.50 & 33.17 & 90.83 \\
 & InternVL-2.5-8B      & 57.50 & 58.00 & \textbf{38.17} & 79.67 \\ \cmidrule(lr){1-6}

\multirow{5}{*}{Diff.} & LLaDA-V       & 54.17 & 51.17 & 33.00 & 86.67 \\
 & LaViDa-LLaDA  & 45.67 & 45.83 & 5.33  & 98.83 \\
 & LaViDa-Dream  & 48.17 & 54.83 & 4.00  & 98.17 \\
 & MMaDA-MixCoT  & 38.50 & 39.67 & 20.83 & 81.00 \\
 & Dream-VL      & 62.17 & \textbf{59.67} & 34.17 & 88.17 \\

\bottomrule
\end{tabular}%
}
\endgroup
\small 
\caption{Comparison of diffusion and autoregressive (AR) vision--language models on the CUB and Dog datasets under controlled option-length settings. Results are grouped by whether class names are included in the answer options. Accuracy (\%) is reported.}
\label{tab:cub_diffusion_vs_ar}
\end{table}

Table~\ref{tab:cub_diffusion_vs_ar} places dLVLMs and AR baselines side by side under identical length-controlled conditions, enabling a direct assessment of the performance gap. On CUB with class names, Qwen2.5-VL-7B achieves 50.70\% accuracy under the \textit{Shorter Correct} condition---more than four times the accuracy of LaViDa-LLaDA (11.50\%) and LaViDa-Dream (9.00\%) under the same condition. InternVL2.5-8B (39.20\%) and LLaDA-V (37.40\%) perform comparably, with LLaDA-V notably outperforming the weaker AR baselines LLaVA-v1.5-7B (23.40\%) and LLaVA-v1.5-13B (19.10\%), suggesting that backbone strength -- rather than generative paradigm alone -- partially determines robustness to length cues. On Dogs with class names, a similar ordering holds: Qwen2.5-VL-7B achieves 58.83\% on \textit{Shorter Correct} versus 21.50--24.50\% for most dLVLMs, with Dream-VL (56.00\%) as the sole diffusion model approaching AR-level robustness. Without class names, the gap widens further for most dLVLMs: LaViDa-Dream drops to 2.80\% on CUB and 4.00\% on Dogs, while InternVL2.5-8B retains 37.80\% and 38.17\% respectively, confirming that the absence of semantic anchors 
disproportionately harms diffusion models.

The \textit{Longer Correct} condition reveals a complementary asymmetry: most dLVLMs exceed 82\% accuracy, and several exceed 90\%, although MMaDA-MixCoT performs notably worse on Dogs (72.17\% with class names and 81.00\% without). These high accuracies show that dLVLMs are not globally incapable of solving the task; rather, their performance improves substantially when the correct answer is longer than the distractors, consistent with a systematic preference for longer options. Notably, LLaDA-V exhibits a less extreme asymmetry, achieving 82.30\% on \textit{Longer Correct} and 47.00\% on \textit{Shorter Correct} for CUB without class names. Taken together, these results show that the length bias documented in the main text is not an artifact of evaluating dLVLMs in isolation. Strong AR models, particularly Qwen2.5-VL-7B and InternVL2.5-8B, generally retain higher \textit{Shorter Correct} accuracy than most dLVLMs, although the gap narrows or reverses in some settings (e.g., LLaDA-V exceeds both AR models on CUB without class names), and most dLVLMs still degrade more severely overall when the correct answer is systematically shorter than its distractors.


\subsection{Linguistic Quality Evaluation Prompt}
\label{sec:linguistic_prompt}

To evaluate the linguistic quality of generated captions, we use GPT-4o-mini as a judge with a structured prompt that defines five error types explicitly. We use deterministic decoding (temperature~=~0) to ensure consistent and reproducible labeling across all 500 captions per model. The prompt instructs the model to evaluate each caption based solely on its text, without assuming access to the image, and to output exactly five binary values in a fixed order. The full prompt is shown in Figure~\ref{fig:linguistic_prompt}.

\begin{figure}[h]
\begin{tcolorbox}[
    colback=gray!8,
    colframe=gray!50,
    title={\textbf{Linguistic Quality Evaluation Prompt (GPT-4o-mini)}},
    fonttitle=\small,
    left=6pt, right=6pt, top=6pt, bottom=6pt
]
\small
You are a linguistic quality evaluator for image captions.

Your task is to evaluate the caption based ONLY on the text itself (do not assume 
access to the image).

For each error type below, output 1 if the error is clearly present, or 0 if it 
is absent. Be conservative: only mark 1 if the issue is obvious.

Output exactly five numbers separated by commas, in this order, with no extra text.

\begin{enumerate}[leftmargin=*, label=\arabic*.]
    \item \textbf{Grammatical error:} Subject-verb disagreement, incorrect tense, 
    missing articles, or malformed sentence structure.
    \item \textbf{Repetition:} Repeated words, phrases, or sentences, or redundant 
    restatement of the same idea.
    \item \textbf{Incoherence:} Sentences do not logically connect, or the caption 
    contradicts itself.
    \item \textbf{Truncation:} Caption ends abruptly or appears incomplete 
    mid-sentence.
    \item \textbf{Unnatural phrasing:} The caption is grammatically correct but 
    sounds awkward, robotic, or not like natural human language.
\end{enumerate}

\textbf{Caption:} \textit{\{caption\}}

\textbf{Answer:}
\end{tcolorbox}
\caption{Prompt used for linguistic quality evaluation of generated captions using 
GPT-4o-mini as judge. The model evaluates each caption across five error types and 
outputs exactly five binary values (0 or 1) separated by commas.}
\label{fig:linguistic_prompt}
\end{figure}

\end{document}